\documentclass[sigconf]{acmart}

\usepackage{algorithm}
\usepackage{algorithmic}
\usepackage{amsthm}
\usepackage{amsmath}
\usepackage{float}
\usepackage{color}
\usepackage{booktabs}
\usepackage{caption}
\usepackage{subcaption}
\usepackage{amsfonts}
\usepackage{multirow}
\usepackage{makecell}

\newcommand{\nop}[1]{}

\newtheorem{lemma0}{Lemma}
\newtheorem{proposition0}{Proposition}
\newtheorem{definition0}{Definition}
\newtheorem{property0}{Property}

\newcolumntype{N}{@{}m{0pt}@{}}

\usepackage{chngcntr,etoolbox}
\newcounter{resetdummycounter}
\newcommand{\resetcounterlist}[1]{%
	\renewcommand*{\do}[1]{\counterwithin*{##1}{resetdummycounter}}%
	\docsvlist{#1}}
\newcommand{\resetcounters}{\stepcounter{resetdummycounter}}
\resetcounterlist{theorem0, lemma0, proposition0}

\nop{
	\fancyhead{}
	\settopmatter{printacmref=false, printfolios=false}
}
\setcopyright{acmlicensed}

\begin{document}
	
	\title{~On Sampling Strategies for Neural Network-based Collaborative Filtering}
	
	\author{Ting Chen}
	\affiliation{%
		\institution{University of California, Los Angeles}
		\city{Los Angeles}
		\state{CA}
		\postcode{90095}
	}
	\email{tingchen@cs.ucla.edu}
	
	\author{Yizhou Sun}
	\affiliation{%
		\institution{University of California, Los Angeles}
		\city{Los Angeles}
		\state{CA}
		\postcode{90095}
	}
	\email{yzsun@cs.ucla.edu}
	
	\author{Yue Shi}
	\authornote{Now at Facebook.}
	\affiliation{%
		\institution{Yahoo! Research}
		\city{Sunnyvale}
		\state{CA}
		\postcode{94089}
	}
	\email{yueshi@acm.org}

	\author{Liangjie Hong}
	\affiliation{%
		\institution{Etsy Inc.}
		\city{Brooklyn}
		\state{NY}
		\postcode{11201}
	}
	\email{lhong@etsy.com}

\begin{abstract}

Recent advances in neural networks have inspired people to design hybrid recommendation algorithms that can incorporate both (1) user-item interaction information and (2) content information including image, audio, and text. Despite their promising results, neural network-based recommendation algorithms pose extensive computational costs, making it challenging to scale and improve upon. In this paper, we  propose a general neural network-based recommendation framework, which subsumes several existing state-of-the-art recommendation algorithms, and address the efficiency issue by investigating sampling strategies in the stochastic gradient descent training for the framework. We tackle this issue by first establishing a connection between the loss functions and the user-item interaction bipartite graph, where the loss function terms are defined on links while major computation burdens are located at nodes. We call this type of loss functions ``graph-based'' loss functions, for which varied mini-batch sampling strategies can have different computational costs. Based on the insight, three novel sampling strategies are proposed, which can significantly improve the training efficiency of the proposed framework (up to $\times 30$ times speedup in our experiments), as well as improving the recommendation performance. Theoretical analysis is also provided for both the computational cost and the convergence. We believe the study of sampling strategies have further implications on general graph-based loss functions, and would also enable more research under the neural network-based recommendation framework.

\end{abstract}

\nop{
%
%

}

\copyrightyear{2017} 
\acmYear{2017} 
\setcopyright{acmlicensed}
\acmConference{KDD '17}{August 13-17, 2017}{Halifax, NS, Canada}\acmPrice{15.00}\acmDOI{10.1145/3097983.3098202}
\acmISBN{978-1-4503-4887-4/17/08}

\maketitle

\section{Introduction}

Collaborative Filtering (CF) has been one of the most effective methods in recommender systems, and methods like matrix factorization \cite{koren2008factorization,koren2009matrix,salakhutdinov2011probabilistic} are widely adopted. However, one of its limitation is the dealing of ``cold-start'' problem, where there are few or no observed interactions for new users or items, such as in news recommendation. To overcome this problem, hybrid methods are proposed to incorporate side information \cite{rendle2010factorization,chen2012svdfeature,singh2008relational}, or item content information \cite{wang2011collaborative,gopalan2014content} into the recommendation algorithm. Although these methods can deal with side information to some extent, they are not effective for extracting features in complicated data, such as image, audio and text. On the contrary, deep neural networks have been shown very powerful at extracting complicated features from those data automatically \cite{krizhevsky2012imagenet,kim2014convolutional}. Hence, it is natural to combine deep learning with traditional collaborative filtering for recommendation tasks, as seen in recent studies \cite{wang2015collaborative,bansal2016ask,zheng2016neural,chen2017text}.

In this work, we generalize several state-of-the-art neural network-based recommendation algorithms \cite{van2013deep,bansal2016ask,chen2017text}, and propose a more general framework that combines both collaborative filtering and deep neural networks in a unified fashion. The framework inherits the best of two worlds: (1) the power of collaborative filtering at capturing user preference via their interaction with items, and (2) that of deep neural networks at automatically extracting high-level features from content data.
However, it also comes with a price. Traditional CF methods, such as sparse matrix factorization \cite{salakhutdinov2011probabilistic,koren2008factorization}, are usually fast to train, while the deep neural networks in general are much more computationally expensive \cite{krizhevsky2012imagenet}. Combining these two models in a new recommendation framework can easily increase computational cost by hundreds of times, thus require a new design of the training algorithm to make it more efficient.

We tackle the computational challenges by first establishing a connection between the loss functions and the user-item interaction bipartite graph. We realize the key issue when combining the CF and deep neural networks are in: the loss function terms are defined over the links, and thus sampling is on links for the stochastic gradient training, while the main computational burdens are located at nodes (e.g., Convolutional Neural Network computation for image of an item). For this type of loss functions, varied mini-batch sampling strategies can lead to different computational costs, depending on how many node computations are required in a mini-batch. The existing stochastic sampling techniques, such as IID sampling, are inefficient, as they do not take into account the node computations that can be potentially shared across links/data points.

Inspired by the connection established, we propose three novel sampling strategies for the general framework that can take coupled computation costs across user-item interactions into consideration. The first strategy is Stratified Sampling, which try to amortize costly node computation by partitioning the links into different groups based on nodes (called stratum), and sample links based on these groups. The second strategy is Negative Sharing, which is based on the observation that interaction/link computation is fast, so once a mini-batch of user-item tuples are sampled, we share the nodes for more links by creating additional negative links between nodes in the same batch. Both strategies have their pros and cons, and to keep their advantages while avoid their weakness, we form the third strategy by combining the above two strategies. Theoretical analysis of computational cost and convergence is also provided. 

Our contributions can be summarized as follows.
\begin{itemize}
	\item We propose a general hybrid recommendation framework (Neural Network-based Collaborative Filtering) combining CF and content-based methods with deep neural networks, which generalize several state-of-the-art approaches.
	\item We establish a connection between the loss functions and the user-item interaction graph, based on which, we propose sampling strategies that can significantly improve training efficiency (up to $\times 30$ times faster in our experiments) as well as the recommendation performance of the proposed framework.
	\item We provide both theoretical analysis and empirical experiments to demonstrate the superiority of the proposed methods.
\end{itemize}
\section{A General Framework for Neural Network-based Collaborative Filtering}
In this section, we propose a general framework for neural network-based Collaborative Filtering that incorporates both interaction and content information.

\subsection{Text Recommendation Problem}
In this work, we use the text recommendation task \cite{wang2011collaborative,wang2015collaborative,bansal2016ask,chen2017text} as an illustrative application for the proposed framework. However, the proposed framework can be applied to more scenarios such as music and video recommendations.

We use $\mathbf{x}_u$ and $\mathbf{x}_v$ to denote features of user $u$ and item $v$, respectively. In text recommendation setting, we set $\mathbf{x}_u$ to one-hot vector indicating $u$'s user id (i.e. a binary vector with only one at the $u$-th position)\footnote{Other user profile features can be included, if available.}, and $\mathbf{x}_v$ as the text sequence, i.e. $\mathbf{x}_v = (w_1, w_2, \cdots, w_t)$. A response matrix $\mathbf{\tilde{R}}$ is used to denote the historical interactions between users and articles, where $\tilde{r}_{uv}$ indicates interaction between a user $u$ and an article $v$, such as ``click-or-not'' and ``like-or-not''. Furthermore, we consider $\mathbf{\tilde{R}}$ as implicit feedback in this work, which means only positive interactions are provided, and non-interactions are treated as negative feedback implicitly.

Given user/item features $\{\mathbf{x}_u \}, \{\mathbf{x}_v\}$ and their historical interaction $\mathbf{\tilde{R}}$, the goal is to learn a model which can rank new articles for an existing user $u$ based on this user's interests and an article's text content.

\subsection{Functional Embedding}
In most of existing matrix factorization techniques \cite{koren2008factorization,koren2009matrix,salakhutdinov2011probabilistic}, each user/item ID is associated with a latent vector $\mathbf{u}$ or $\mathbf{v}$ (i.e., embedding), which can be considered as a simple linear transformation from the one-hot vector represented by their IDs, i.e. $\mathbf{u}_u = \mathbf{f}(\mathbf{x}_u) = \mathbf{W}^T \mathbf{x}_u$ ($\mathbf{W}$ is the embedding/weight matrix). Although simple, this direct association of user/item ID with representation make it less flexible and unable to incorporate features such as text and image.

In order to effectively incorporate user and item features such as content information, it has been proposed to replace embedding vectors $\mathbf{u}$ or $\mathbf{v}$ with functions such as decision trees \cite{zhou2011functional} and some specific neural networks \cite{bansal2016ask,chen2017text}. Generalizing the existing work, we propose to replace the original embedding vectors $\mathbf{u}$ and $\mathbf{v}$ with general differentiable functions $\mathbf{f}(\cdot) \in \mathbb{R}^d$ and $\mathbf{g}(\cdot) \in \mathbb{R}^d$ that take user/item features $\mathbf{x}_u, \mathbf{x}_v$ as their inputs. Since the user/item embeddings are the output vectors of functions, we call this approach \textit{Functional Embedding}. After embeddings are computed, a score function $r(u,v)$ can be defined based on these embeddings for a user/item pair $(u,v)$, such as vector dot product $r(u, v) = \mathbf{f}(\mathbf{x}_u)^T \mathbf{g}(\mathbf{x}_v)  $ (used in this work), or a general neural network.
\begin{figure}[t!]
	\centering
	\includegraphics[width=0.34\textwidth]{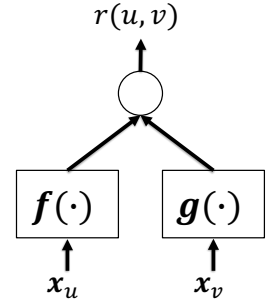}
	\caption{The functional embedding framework.}
	\label{fig:framework}
\end{figure}
The model framework is shown in Figure \ref{fig:framework}. It is easy to see that our framework is very general, as it does not explicitly specify the feature extraction functions, as long as the functions are differentiable. In practice, these function can be specified with neural networks such as CNN or RNN, for extracting high-level information from image, audio, or text sequence. When there are no features associated, it degenerates to conventional matrix factorization where user/item IDs are used as their features.

For simplicity, we will denote the output of $\mathbf{f}(\mathbf{x}_u)$ and $\mathbf{g}(\mathbf{x}_v)$ by $\mathbf{f}_u$ and $\mathbf{g}_v$, which are the embedding vectors for user $u$ and item $v$.

\subsection{Loss Functions for Implicit Feedback}
In many real-world applications, users only provide positive signals according to their preferences, while negative signals are usually implicit. This is usually referred as ``implicit feedback'' \cite{pan2008one,hu2008collaborative,rendle2009bpr}. In this work, we consider two types of loss functions that can handle recommendation tasks with implicit feedback, namely, pointwise loss functions and pairwise loss functions.
\begin{table}[!t]
	\centering
	\small{
	\caption{Examples of loss functions for recommendation.}
	\label{tab:loss_inst}
	\begin{tabular}{c}
		\hline
		{Pointwise loss}\\
		\hline
		SG-loss  \cite{mikolov2013distributed}:  -$\sum_{(u, v) \in \mathcal{D}} \bigg(\log\sigma(\mathbf{f}_u^T \mathbf{g}_v) + \lambda \mathbb{E}_{v'\sim P_n} \log\sigma(-\mathbf{f}_u^T \mathbf{g}_{v'})\bigg)$ \\  
		MSE-loss  \cite{van2013deep}: $ \sum_{(u, v) \in \mathcal{D}} \bigg((\tilde{r}^+_{uv} - \mathbf{f}_u^T \mathbf{g}_v)^2 + \lambda \mathbb{E}_{v' \sim P_n} (\tilde{r}^-_{uv'} - \mathbf{f}_u^T \mathbf{g}_{v'})^2\bigg)$  \\ \hline
		{Pairwise loss}\\
		\hline
		Log-loss \cite{rendle2009bpr}: -$\sum_{(u, v) \in \mathcal{D}} \mathbb{E}_{v'\sim P_n} \log \sigma\bigg(\gamma(\mathbf{f}_u^T \mathbf{g}_v - \mathbf{f}_u^T \mathbf{g}_{v'})\bigg)$\\ 
		Hinge-loss \cite{weimer2008improving}:
		$\sum_{(u, v) \in \mathcal{D}} \mathbb{E}_{{v'}\sim P_n} \max\bigg(\mathbf{f}_u^T \mathbf{g}_{v'} - \mathbf{f}_u^T \mathbf{g}_v + \gamma, 0\bigg)$ \\
		\hline
	\end{tabular}
	}
\end{table}
Pointwise loss functions have been applied to such problems in many existing work. In \cite{van2013deep,wang2015collaborative,bansal2016ask}, mean square loss (MSE) has been applied where ``negative terms'' are weighted less. And skip-gram (SG) loss has been successfully utilized to learn robust word embedding \cite{mikolov2013distributed}.

These two loss functions are summarized in Table \ref{tab:loss_inst}. Note that we use a weighted expectation term over all negative samples, which can be approximated with small number of samples. We can also abstract the pointwise loss functions into the following form:
\begin{equation}
\label{eq:point}
\begin{split}
\mathcal{L}_{\text{pointwise}} = \mathbb{E}_{u\sim P_d(u)}\bigg[
&\mathbb{E}_{v\sim P_d(v|u)} c^+_{uv} \mathcal{L^+}(u, v|\theta) \\
& +\mathbb{E}_{v'\sim P_n(v')} c^-_{uv'} \mathcal{L^-}(u, v'|\theta) 
\bigg]
\end{split}
\end{equation}
where $P_d$ is (empirical) data distribution, $P_n$ is user-defined negative data distribution, $c$ is user defined weights for the different user-item pairs, $\theta$ denotes the set of all parameters, $\mathcal{L}^+(u, v|\theta)$ denotes the loss function on a single positive pair $(u,v)$, and $\mathcal{L}^-(u, v|\theta)$ denotes the loss on a single negative pair. Generally speaking, given a user $u$, pointwise loss function encourages her score with positive items $\{v\}$, and discourage her score with negative items $\{v'\}$.

When it comes to ranking problem as commonly seen in implicit feedback setting, some have argued that the pairwise loss would be advantageous \cite{rendle2009bpr,weimer2008improving}, as pairwise loss encourages ranking of positive items above negative items for the given user. Different from pointwise counterparts, pairwise loss functions are defined on a triplet of $(u, v, v')$, where $v$ is a positive item and $v'$ is a negative item to the user $u$. Table \ref{tab:loss_inst} also gives two instances of such loss functions used in existing papers \cite{rendle2009bpr,weimer2008improving} (with $\gamma$ being the pre-defined ``margin'' parameter). We can also abstract pairwise loss functions by the following form:
\begin{equation}
\label{eq:pair}
\begin{split}
\mathcal{L}_{\text{pairwise}} = \mathbb{E}_{u\sim P_d(u)} \mathbb{E}_{v\sim P_d(v|u)}\mathbb{E}_{v'\sim P_n(v')} c_{uvv'} \mathcal{L}(u, v, v'|\theta)
\end{split}
\end{equation}
where the notations are similarly defined as in Eq. \ref{eq:point} and $\mathcal{L}(u, v, v'|\theta)$ denotes the loss function on the triplet $(u,v,v')$.

\subsection{Stochastic Gradient Descent Training and Computational Challenges}
\begin{figure}[h]
	\centering
	\includegraphics[width=0.3\textwidth]{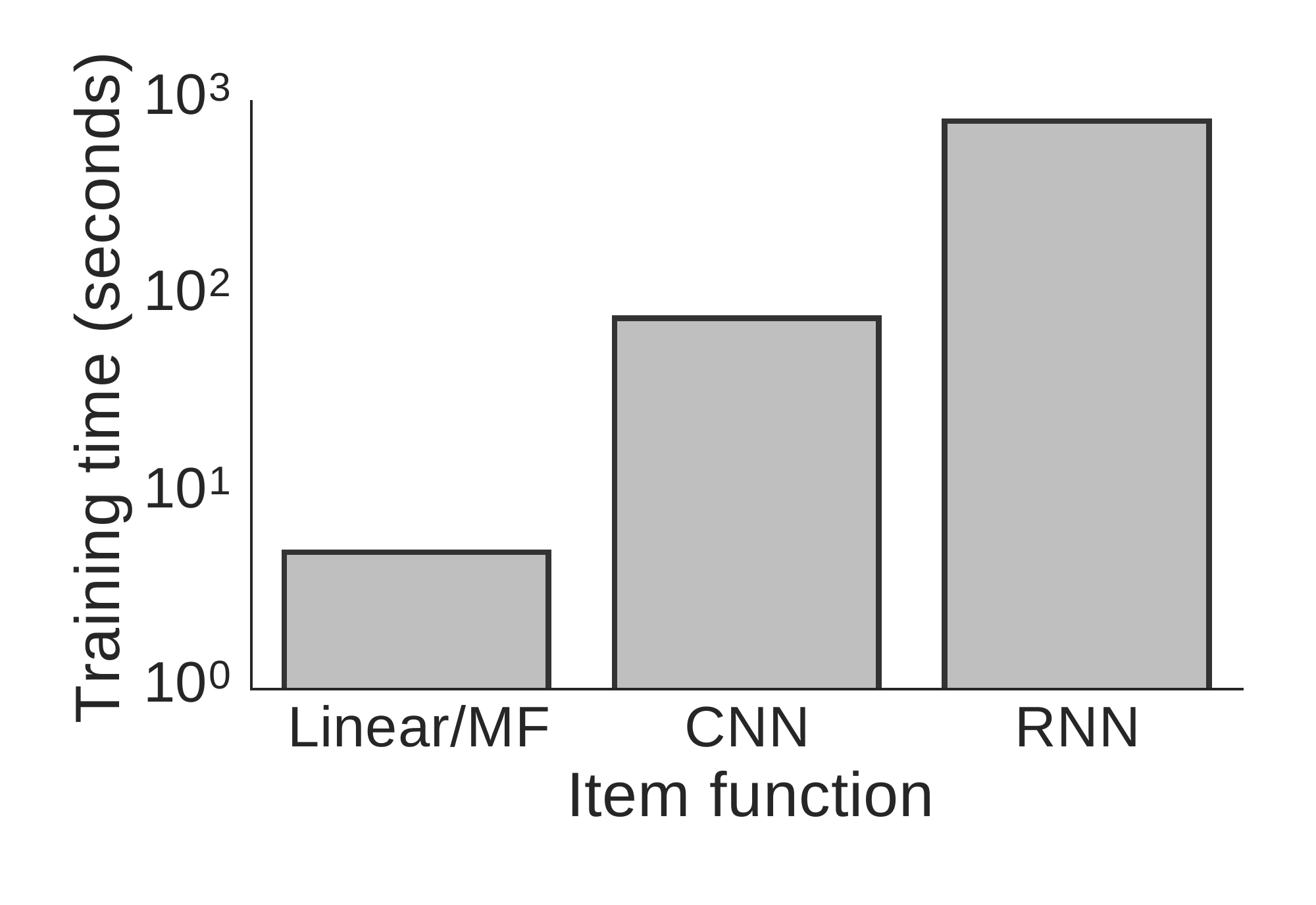}
	\caption{Model training time per epoch with different types of item functions (in log-scale).}
	\label{fig:operational}
\end{figure}

To train the model, we use stochastic gradient descent based algorithms \cite{bottou2010large,kingma2014adam}, which are widely used for training matrix factorization and neural networks. The main flow of the training algorithm is summarized in Algorithm \ref{alg:original}.
\begin{algorithm}[!t]
	\caption{Standard model training procedure}
	\label{alg:original}
	\begin{algorithmic}
		\WHILE{not converged}
		\STATE // mini-batch sampling
		\STATE draw a mini-batch of user-item tuples $(u, v)$\footnotemark
		\STATE // forward pass
		\STATE compute $\mathbf{f}(\mathbf{x}_u)$, $\mathbf{g}(\mathbf{x}_v)$ and their interaction $\mathbf{f}_u^T \mathbf{g}_v$
		\STATE compute the loss function $\mathcal{L}$
		\STATE // backward pass
		\STATE compute gradients and apply SGD updates
		\ENDWHILE
	\end{algorithmic}
\end{algorithm}
\footnotetext{Draw a mini-batch of user-item triplets $(u, v, v')$ if a pairwise loss function is adopted.}
By adopting the functional embedding with (deep) neural networks, we can increase the power of the model, but it also comes with a cost. Figure \ref{fig:operational} shows the training time (for CiteULike data) with different item functions $\mathbf{g}(\cdot)$, namely linear embedding taking item id as feature (equivalent to conventional MF), CNN-based content embedding, and RNN/LSTM-based content embedding. We see orders of magnitude increase of training time for the latter two embedding functions, which may create barriers to adopt models under this framework.

Breaking down the computation cost of the framework, there are three major parts of computational cost. The first part is the user based computation (denoted by $t_f$ time units per user), which includes forward computation of user function $\mathbf{f}(\mathbf{x}_u)$, and backward computation of the function output w.r.t. its parameters. The second part is the item based computation (denoted by $t_g$ time units per item), which similarly includes forward computation of item function $\mathbf{g}(\mathbf{x}_v)$, as well as the back computation. The third part is the computation for interaction function (denoted by $t_i$ time units per interaction). The total computational cost for a mini-batch is then $t_f \times \text{\# of users} + t_g \times \text{\# of items} + t_i \times \text{\# of interactions}$, with some other minor operations which we assume ignorable. In the text recommendation application, user IDs are used as user features (which can be seen as linear layer on top of the one-hot inputs), (deep) neural networks are used for text sequences, vector dot product is used as interaction function, thus the dominant computational cost is $t_g$ (orders of magnitude larger than $t_f$ and $t_i$). In other words, we assume $t_g \gg t_f, t_i$ in this work.

\section{Mini-Batch Sampling Strategies For Efficient Model Training}
In this section, we propose and discuss different sampling strategies that can improve the efficiency of the model training.

\subsection{Computational Cost in a Graph View}
\begin{figure}[t!]
	\centering
	\includegraphics[width=0.35\textwidth]{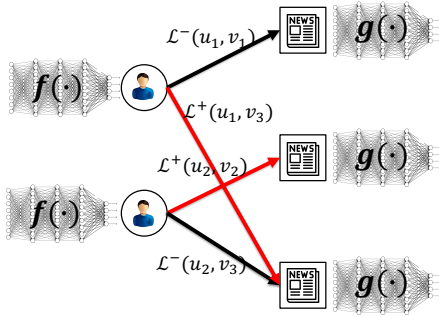}
	\caption{The bipartite interaction graph for pointwise loss functions, where loss functions are defined over links. The pairwise loss functions are defined over pairs of links.}
	\label{fig:connection}
\end{figure}

Before the discussion of different sampling strategies, we motivate our readers by first making a connection between the loss functions and the bipartite graph of user-item interactions. In the loss functions laid out before, we observed that each loss function term in Eq. \ref{eq:point}, namely, $\mathcal{L}(u, v)$, involves a pair of user and item, which corresponds to a link in their interaction graph. And two types of links corresponding to two types of loss terms in the loss functions, i.e., positive links/terms and negative links/terms. Similar analysis holds for pairwise loss in Eq. \ref{eq:pair}, though there are slight differences as each single loss function corresponds to a pair of links with opposite signs on the graph. We can also establish a correspondence between user/item functions and nodes in the graph, i.e., $\mathbf{f}(u)$ to user node $u$ and $\mathbf{g}(v)$ to item node $v$. The connection is illustrated in Figure \ref{fig:connection}. Since the loss functions are defined over the links, we name them ``\textit{graph-based}'' loss functions to emphasize the connection.

The key observation for graph-based loss functions is that: the loss functions are defined over links, but the major computational burden are located at nodes (due to the use of costly $\mathbf{g}(\cdot)$ function). Since each node is associated with multiple links, which are corresponding to multiple loss function terms, the computational costs of loss functions over links are coupled (as they may share the same nodes) when using mini-batch based SGD. Hence, varied sampling strategies yield different computational costs. For example, when we put links connected to the same node together in a mini-batch, the computational cost can be lowered as there are fewer $\mathbf{g}(\cdot)$ to compute\footnote{This holds for both forward and backward computation. For the latter, the gradient from different links can be aggregated before back-propagating to $\mathbf{g}(\cdot)$.}. This is in great contrast to conventional optimization problems, where each loss function term dose not couple with others  in terms of computation cost.
\begin{figure*}[t!]
	\centering
	\begin{subfigure}[b]{0.18\textwidth}
		\includegraphics[width=\textwidth]{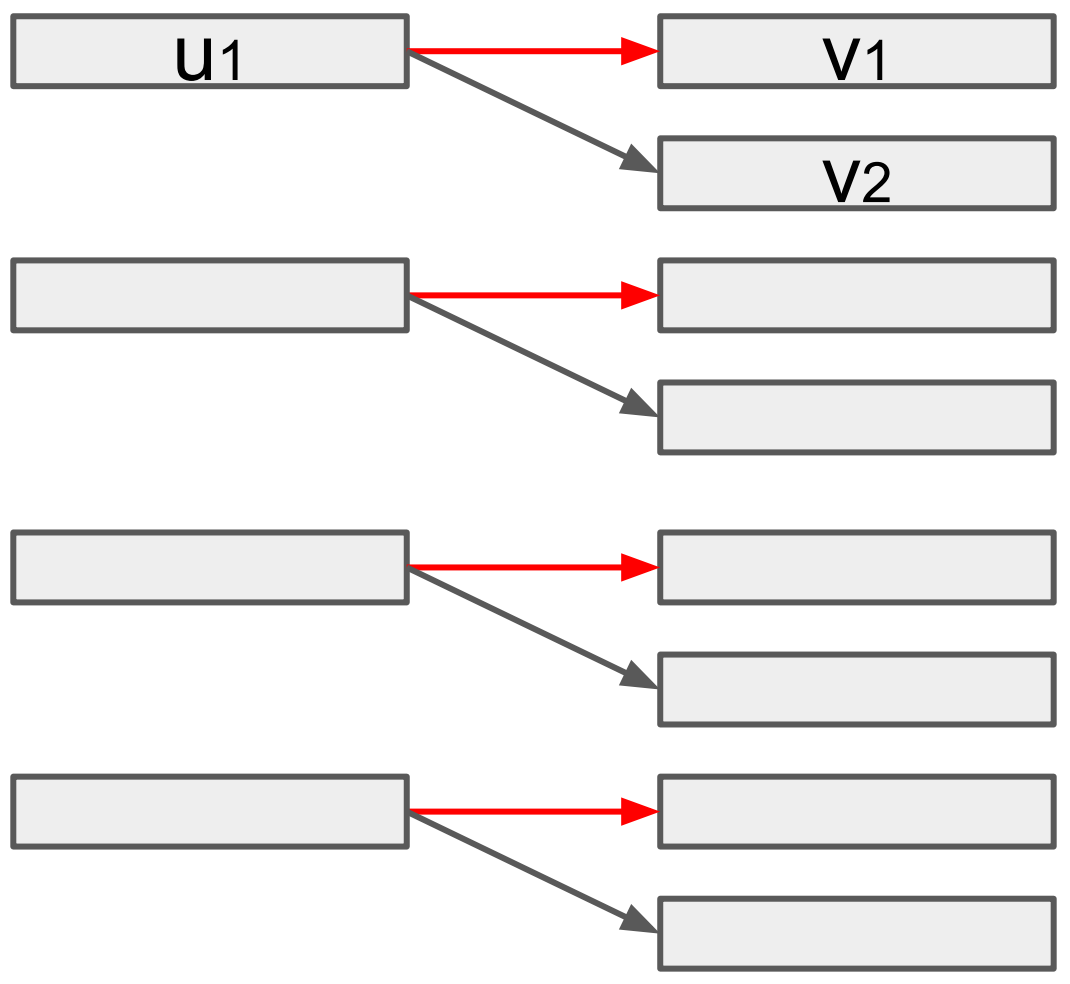}
		\caption{Negative}
		\label{fig:sampling_a}
	\end{subfigure}\hspace{1.5em}
	\begin{subfigure}[b]{0.18\textwidth}
		\includegraphics[width=\textwidth]{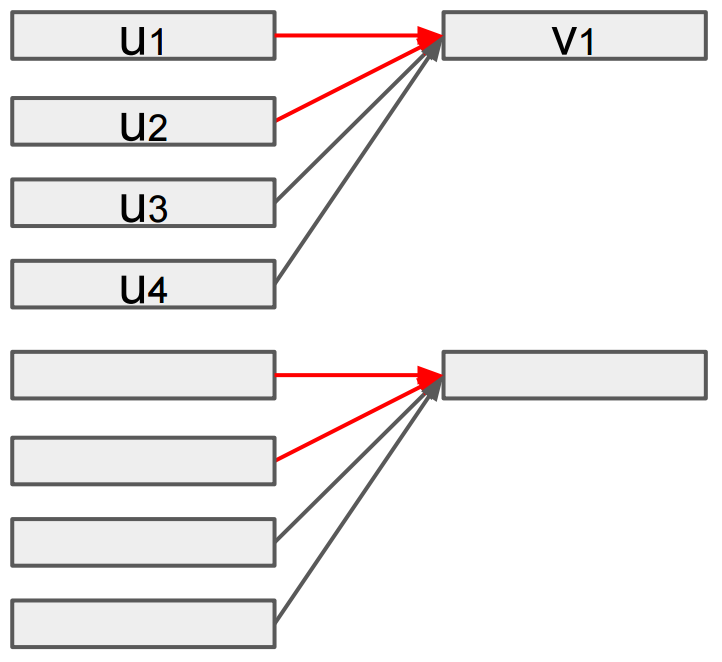}
		\caption{Stratified (by Items)}
		\label{fig:sampling_b}
	\end{subfigure}\hspace{1.5em}
	\begin{subfigure}[b]{0.18\textwidth}
		\includegraphics[width=\textwidth]{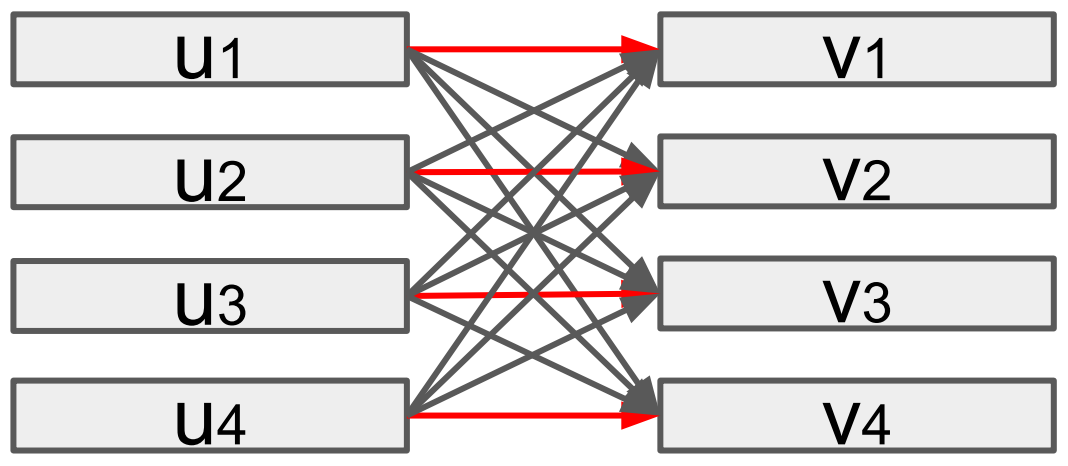}
		\vspace{1.5em}
		\caption{Negative Sharing}
		\label{fig:sampling_c}
	\end{subfigure}\hspace{1.5em}
	\begin{subfigure}[b]{0.18\textwidth}
		\includegraphics[width=\textwidth]{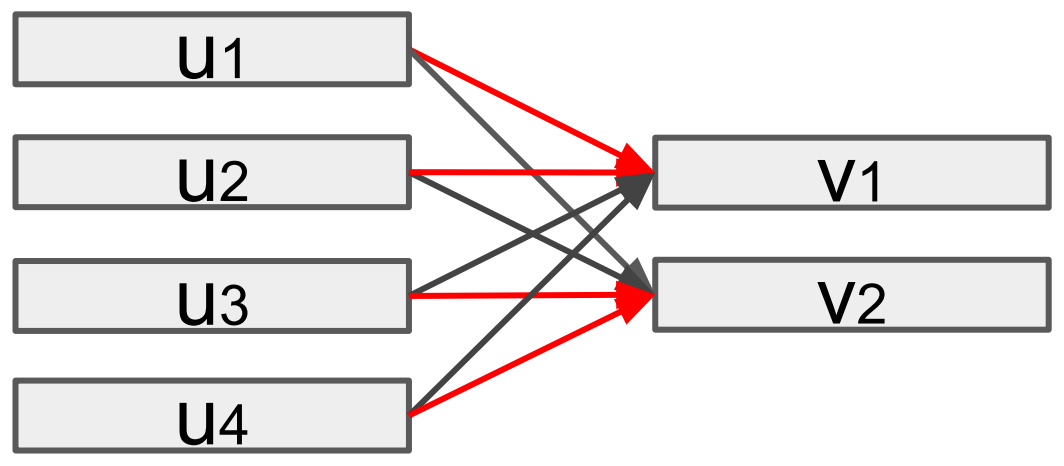}
		\vspace{1.5em}
		\caption{Stratified with N.S.}
		\label{fig:sampling_d}
	\end{subfigure}
	\caption{Illustration of four different sampling strategies. \ref{fig:sampling_b}-\ref{fig:sampling_d} are the proposed sampling strategies. Red lines denote positive links/interactions, and black lines denote negative links/interactions.}
	\label{fig:sampling}
\end{figure*}

\subsection{Existing Mini-Batch Sampling Strategies}
In standard SGD sampler, (positive) data samples are drawn uniformly at random for gradient computation. Due to the appearance of negative samples, we draw negative samples from some predefined probability distribution, i.e.  $(u', v')\sim P_n(u', v')$. We call this approach ``\textit{IID Sampling}'', since each positive link is dependently and identical distributed, and the same holds for negative links (with a different distribution).

Many existing algorithms with graph-based loss functions \cite{mikolov2013distributed,tang2015line,bansal2016ask} adopt the ``\textit{Negative Sampling}'' strategy, in which $k$ negative samples are drawn whenever a positive example is drawn. The negative samples are sampled based on the positive ones by replacing the items in the positive samples. This is illustrated in Algorithm \ref{alg:neg_sampling} and Figure \ref{fig:sampling_a}.

\begin{algorithm}[h]
	\caption{Negative Sampling \cite{mikolov2013efficient,tang2015line,bansal2016ask}}
	\label{alg:neg_sampling}
	\begin{algorithmic}
		\STATE \algorithmicrequire~number of positive links in a mini-batch $b$, number of negative links per positive one: $k$
		\STATE draw $b$ positive links uniformly at random
		\FOR{each of $b$ positive links}
		\STATE draw $k$ negative links by replacing true item $v$ with $v' \propto P_n(v')$
		\ENDFOR
	\end{algorithmic}
\end{algorithm}

The IID Sampling strategy dose not take into account the property of graph-based loss functions, since samples are completely independent of each other. Hence, the computational cost in a single mini-batch cannot be amortized across different samples, leading to very extensive computations with (deep) neural networks. The Negative Sampling does not really help, since the item function computation cost $t_g$ is the dominant one. To be more specific, consider a mini-batch with $b(1+k)$ links sampled by IID Sampling or Negative Sampling, we have to conduct item based $\mathbf{g}(\cdot)$ computation $b(1 + k)$ times, since items in a mini-batch are likely to be non-overlapping with sufficient large item sets.

\subsection{The Proposed Sampling Strategies}

\subsubsection{Stratified Sampling (by Items)}
Motivated by the connection between the loss functions and the bipartite interaction graph as shown in Figure \ref{fig:connection}, we propose to sample links that share nodes, in particular those with high computational cost (i.e. $t_g$ for item function $\mathbf{g}(\cdot)$ in our case). By doing so, the computational cost within a mini-batch can be amortized, since fewer costly functions are computed (in both forward and backward propagations).

In order to achieve this, we (conceptually) partition the links, which correspond to loss function terms, into \textit{strata}. A \textit{stratum} in the strata is a set of links on the bipartite graph sharing the same source or destination node. Instead of drawing links directly for training, we will first draw stratum and then draw both positive and negative links. Since we want each stratum to share the same item, we can directly draw an item and then sample its links. The details are given in Algorithm \ref{alg:stratified_sampling} and illustrated in Figure \ref{fig:sampling_b}.

\begin{algorithm}[t!]
	\caption{Stratified Sampling (by Items)}
	\label{alg:stratified_sampling}
	\begin{algorithmic}
		\STATE \algorithmicrequire~number of positive links in a mini-batch: $b$, number of positive links per stratum: $s$, number of negative links per positive one: $k$
		\REPEAT
		\STATE draw an item $v \propto P_d(v)$
		\STATE draw $s$ positive users $\{u\}$ of $v$ uniformly at random
		\STATE draw $k\times s$ negative users $ \{u'\} \propto P_d(u')$
		\UNTIL{a mini-batch of $b$ positive links are sampled}
	\end{algorithmic}
\end{algorithm}

Compared to Negative Sampling in Algorithm \ref{alg:neg_sampling}, there are several differences: (1) Stratified Sampling can be based on either item or user, but in the negative sampling only negative items are drawn; and (2) each node in stratified sampling can be associated with more than 1 positive link (i.e., $s>1$, which can help improve the speedup as shown below), while in negative sampling each node is only associated with one positive link.

Now we consider its speedup for a mini-batch including $b$ positive links/interactions and $bk$ negative ones, which contains $b(1+k)$ users and $b/s$ items. The Stratified Sampling (by Items) only requires $b/s$ computations of $\mathbf{g}(\cdot)$ functions, while the Negative Sampling requires $b(1 + k)$ computations. Assuming $t_g \gg t_f, t_i$, i.e. the computation cost is dominated by the item function $g(\cdot)$, the Stratified Sampling (by Items) can provide $s(1 + k)$ times speedup in a mini-batch. With $s=4, k=10$ as used in some of our experiments, it yields to $\times 40$ speedup optimally. However, it is worth pointing out that item-based Stratified Sampling cannot be applied to pairwise loss functions, which compare preferences over items based on a given user.

\subsubsection{Negative Sharing}

The idea of Negative Sharing is inspired from a different aspect of the connection between the loss functions and the bipartite interaction graph. Since $t_i \ll t_g$, i.e. the computational cost of interaction function (dot product) is ignorable compared to that of item function, when a mini-batch of users and items are sampled, increasing the number of interactions among them may not result in a significant increase of computational cost. This can be achieved by creating a complete bipartite graph for a mini-batch by adding negative links between all non-interaction pairs between users and items. Using this strategy, we can draw NO negative links at all!

More specifically, consider the IID Sampling, when $b$ positive links are sampled, there will be $b$ users and $b$ items involved (assuming the sizes of user set and item set are much larger than $b$). Note that, there are $b(b-1)$ non-interactions in the mini-batch, which are not considered in IID Sampling or Negative Sampling, instead they draw additional negative samples. Since the main computational cost of training is on the node computation and the node set is fixed given the batch of $b$ positive links, we can share the nodes for negative links without increasing much of computational burdens. Based on this idea, Algorithm \ref{alg:neg_shared} summarizes an extremely simple sampling procedure, and it is illustrated in Figure \ref{fig:sampling_c}.

\begin{algorithm}[t!]
	\caption{Negative Sharing}
	\label{alg:neg_shared}
	\begin{algorithmic}
		\STATE \algorithmicrequire~number of positive links in a mini-batch: $b$
		\STATE draw $b$ positive user-item pairs $\{(u, v)\}$ uniformly at random
		\STATE construct negative pairs by connecting non-linked users and items in the batch
	\end{algorithmic}
\end{algorithm}

Since Negative Sharing avoids sampling $k$ negative links, it only contains $b$ items while in Negative Sampling contains $b(1+k)$ items. So it can provide $(1 + k)$ times speedup compared to Negative Sampling (assuming $t_g \gg t_f, t_i$, and total interaction cost is still insignificant). Given the batch size $b$ is usually larger than $k$ (e.g., $b=512, k=20$ in our experiments), much more negative links (e.g. $512 \times 511$) will also be considered, this is helpful for both faster convergence and better performance, which is shown in our experiments. However, as the number of negative samples increases, the performance and the convergence will not be improved linearly. diminishing return is expected.

\subsubsection{Stratified Sampling with Negative Sharing}
The two strategies above can both reduce the computational cost by smarter sampling of the mini-batch. However, they both have weakness: Stratified Sampling cannot deal with pairwise loss and it is still dependent on the number of negative examples $k$, and Negative Sharing introduces a lot of negative samples which may be unnecessary due to diminishing return.

The good news is, the two sampling strategies are proposed from different perspectives, and combining them together can preserve their advantages while avoid their weakness. This leads to the Stratified Sampling with Negative Sharing, which can be applied to both pointwise and pairwise loss functions, and it can have flexible ratio between positive and negative samples (i.e. more positive links given the same negative links compared to Negative Sharing). To do so, basically we sample positive links according to Stratified Sampling, and then sample/create negative links by treating non-interactions as negative links. The details are given in Algorithm \ref{alg:stratified_sampling_with_neg_shared} and illustrated in Figure \ref{fig:sampling_d}.

\begin{algorithm}[t!]
	\caption{Stratified Sampling with Negative Sharing}
	\label{alg:stratified_sampling_with_neg_shared}
	\begin{algorithmic}
		\STATE \algorithmicrequire~number of positive links in a mini-batch: $b$, number of positive links per stratum: $s$
		\REPEAT
		\STATE draw an item $v \propto P_d(v)$
		\STATE draw $s$ positive users of item $v$ uniformly at random
		\UNTIL{a mini-batch of $b/s$ items are sampled}
		\STATE construct negative pairs by connecting non-linked users and items in the batch
	\end{algorithmic}
\end{algorithm}

Computationally, Stratified Sampling with Negative Sharing only involve $b/s$ item nodes in a mini-batch, so it can provide the same $s(1 + k)$ times speedup over Negative Sampling as Stratified Sampling (by Items) does, but it will utilize much more negative links compared to Negative Sampling. For example, in our experiments with $b=512, s=4$, we have $127$ negative links per positive one, much larger than $k=10$ in Negative Sampling, and only requires $1/4$ times of $\mathbf{g}(\cdot)$ computations compared to Negative Sharing.

\begin{table*}[!h]
	\centering
	\caption{Computational cost analysis for a batch of $b$ positive links. We use vec to denote vector multiplication, and mat to denote matrix multiplication. Since $t_g \gg t_f, t_i$ in practice, the theoretical speedup per iteration can be approximated by comparing the number of $t_g$ computation, which is colored red below. The number of iterations to reach a referenced loss is related to the number of negative links in each mini-batch.}
	\label{tab:cost_model}
	\begin{tabular}{cccccccc}
		\Xhline{2.5\arrayrulewidth}
		\multicolumn{1}{c}{Sampling}  & \multicolumn{1}{c}{\# pos. links} & \multicolumn{1}{c}{\# neg. links} & \multicolumn{1}{c}{\# $t_f$} & \multicolumn{1}{c}{\textcolor{red}{\# $t_g$}} & \multicolumn{1}{c}{\# $t_i$} & \multicolumn{1}{c}{pointwise} & \multicolumn{1}{c}{pairwise}\\ \hline
		IID \cite{bottou2010large}         & $b $                                    & $bk$                                    & $b(1+k)$                                    & \textcolor{red}{$b(1+k)$}                             & $b(1+k)$ vec                         & \checkmark & $\times$ \\
		Negative \cite{mikolov2013efficient,tang2015line,bansal2016ask}          & $b $                                    & $bk$                                    & $b$                                    & \textcolor{red}{$b(1+k)$}                              & $b(1+k)$ vec                         & \checkmark & \checkmark\\ \hline
		Stratified (by Items) & $b$                                     & $bk$                                    & $b(1+k)$                               & \textcolor{red}{$\frac{b}{s}$}                                  & $b(1+k)$ vec                        & \checkmark & $\times$ \\
		Negative Sharing              & $b$                                     & $b(b-1)$                                & $b $                                   & \textcolor{red}{$b$}                 & $b\times b$ mat                         & \checkmark & \checkmark \\
		Stratified with N.S.   & $b $                                    & $ \frac{b(b-1)}{s}$                              & $b$                                    & \textcolor{red}{$\frac{b}{s}$}                                  & $b\times\frac{b}{s}$ mat                    &\checkmark & \checkmark  \\
		\Xhline{2.5\arrayrulewidth}
	\end{tabular}
\end{table*}

\subsubsection{Implementation Details}

When the negative/noise distribution $P_n$ is not unigram\footnote{Unigram means proportional to item frequency, such as node degree in user-item interaction graph.}, we need to adjust the loss function in order to make sure the stochastic gradient is unbiased. For pointwise loss, each of the negative term is adjusted by multiplying a weight of $\frac{P_n(v')}{P_d(v')}$; for pairwise loss, each term based on a triplet of $(u, v, v')$ is adjusted by multiplying a weight of $\frac{P_n(v')}{P_d(v')}$ where $v'$ is the sampled negative item.

Instead of sampling, we prefer to use shuffling as much as we can, which produces unbiased samples while yielding zero variance. This can be a useful trick for achieving better performance when the number of drawn samples are not large enough for each loss terms. For IID and Negative Sampling, this can be easily done for positive links by simply shuffling them. As for the Stratified Sampling (w./wo. Negative Sharing), instead of shuffling the positive links directly, we shuffle the randomly formed strata (where each stratum contains roughly a single item)\footnote{This can be done by first shuffling users associated with each item, and then concatenating all links according to items in random order, random strata is then formed by segmenting the list.}. All other necessary sampling operations required are sampling from discrete distributions, which can be done in $O(1)$ with Alias method.

In Negative Sharing (w./wo. Stratified Sampling), We can compute the user-item interactions with more efficient operator, i.e. replacing the vector dot product between each pair of $(\mathbf{f}, \mathbf{g})$ with matrix multiplication between $(\mathbf{F}, \mathbf{G})$, where $\mathbf{F} = [\mathbf{f}_{u_1}, \cdots, \mathbf{f}_{u_n}]$, $\mathbf{G} = [\mathbf{g}_{v_1}, \cdots, \mathbf{g}_{v_m}]$. Since matrix multiplication is higher in BLAS level than vector multiplication \cite{ji2016parallelizing}, even we increase the number of interactions, with medium matrix size (e.g. 1000$\times$ 1000) it does not affect the computational cost much in practice.

\subsection{Computational Cost and Convergence Analysis}

Here we provide a summary for the computational cost for different sampling strategies discussed above, and also analyze their convergences. Two aspects that can lead to speedup are analyzed: (1) the computational cost for a mini-batch, i.e. per iteration, and (2) the number of iterations required to reach some referenced loss. 

\subsubsection{Computational Cost}

To fairly compare different sampling strategies, we fix the same number of positive links in each of the mini-batch, which correspond to the positive terms in the loss function. Table \ref{tab:cost_model} shows the computational cost of different sampling strategies for a given mini-batch. Since $t_g \gg t_f,t_i$ in practice, we approximate the theoretical speedup per iteration by comparing the number of $t_g$ computation. We can see that the proposed sampling strategies can provide $(1+k)$, by Negative Sharing, or $s(1+k)$, by Stratified Sampling (w./w.o. Negative Sharing), times speedup for each iteration compared to IID Sampling or Negative Sampling. As for the number of iterations to reach a reference loss, it is related to number of negative samples utilized, which is analyzed below.

\subsubsection{Convergence Analysis}

We want to make sure the SGD training under the proposed sampling strategies can converge correctly. The necessary condition for this to hold is the stochastic gradient estimator has to be unbiased, which leads us to the following lemma.

\begin{lemma0}
	\label{th:convergence_lemma} (unbiased stochastic gradient)
	Under sampling Algorithm \ref{alg:neg_sampling}, \ref{alg:stratified_sampling}, \ref{alg:neg_shared}, and \ref{alg:stratified_sampling_with_neg_shared}, we have $\mathbb{E}_B[\nabla \mathcal{L}_{B}(\theta^t)] = \nabla\mathcal{L}(\theta^t)$. In other words, the stochastic mini-batch gradient equals to true gradient in expectation.
\end{lemma0}

This holds for both pointwise loss and pairwise loss. It is guaranteed since we draw samples stochastically and re-weight certain samples accordingly. The detailed proof can be found in the supplementary material.

Given this lemma, we can further analyze the convergence behavior of the proposed sampling behaviors. Due to the highly non-linear and non-convex functions composed by (deep) neural networks, the convergence rate is usually difficult to analyze. So we show the SGD with the proposed sampling strategies follow a local convergence bound (similar to \cite{ghadimi2013stochastic,reddi2016stochastic}).

\begin{proposition0}\label{th:convergence} (local convergence)
	Suppose $\mathcal{L}$ has $\sigma$-bounded gradient; let $\eta_t = \eta = c/ \sqrt{T}$ where $c = \sqrt{\frac{2(\mathcal{L}(\theta^0) - \mathcal{L}(\theta^*)}{L \sigma^2}}$, and $\theta^*$ is the minimizer to $\mathcal{L}$. Then, the following holds for the proposed sampling strategies given in Algorithm \ref{alg:neg_sampling}, \ref{alg:stratified_sampling}, \ref{alg:neg_shared}, \ref{alg:stratified_sampling_with_neg_shared}
	
	$$
	\min_{0\le t\le T-1} \mathbb{E}[\|\nabla \mathcal{L}(\theta^t)\|^2] \le  \sqrt{\frac{2(\mathcal{L}(\theta^0) - \mathcal{L}(\theta^*))}{T}} \sigma
	$$
\end{proposition0}

The detailed proof is also given in the supplementary material.

Furthermore, utilizing more negative links in each mini-batch can lower the expected stochastic gradient variance. As shown in \cite{zhao2014accelerating,zhao2015stochastic}, the reduction of variance can lead to faster convergence. This suggests that Negative Sharing (w./wo. Stratified Sampling) has better convergence than the Stratified Sampling (by Items).

\section{Experiments}

\subsection{Data Sets}

Two real-world text recommendation data sets are used for the experiments. The first data set CiteULike, collected from CiteULike.org, is provided in~\cite{wang2011collaborative}. The CiteULike data set contains users bookmarking papers, where each paper is associated with a title and an abstract. The second data set is a random subset of Yahoo! News data set \footnote{https://webscope.sandbox.yahoo.com/catalog.php?datatype=r\&did=75}, which contains users clicking on news presented at Yahoo!. There are 5,551 users and 16,980 items, and total of 204,986 positive interactions in CiteULike data. As for Yahoo! News data, there are 10,000 users, 58,579 items and 515,503 interactions.

Following \cite{chen2017text}, we select a portion (20\%) of items to form the pool of test items. All user interactions with those test items are held-out during training, only the remaining user-item interactions are used as training data, which simulates the scenarios for recommending newly-emerged text articles.

\nop{
\begin{table}[t!]
	\small
	\centering
\caption{\label{tab:data_stat1} Data statistics for user, items and their interactions.}
\begin{tabular}{lrrr}
	\toprule
	{} &  \# of user &  \# of item &  \# of interaction \\
	\midrule
	Citeulike         &       5,551 &      16,980 &            204,986 \\
	News             &      10,000 &      58,579 &            515,503 \\
	\bottomrule
\end{tabular}
\end{table}

\begin{table}[t!]
\small
\centering
\caption{\label{tab:data_stat2} Data statistics for text content.}
\begin{tabular}{lrrrrr}
	\toprule
	{} &  voc. size & max & min & mean & median \\
	\midrule
	Citeulike  &  23,011 &         300 &           22 &          194 &            186 \\
	News    &            41,537 &         200 &           2 &           89 &             90 \\
	\bottomrule
\end{tabular}
\end{table}

The detailed data set statistics are shown in Table~\ref{tab:data_stat1} and~\ref{tab:data_stat2}.
}
\subsection{Experimental Settings}

The main purpose of experiments is to compare the efficiency and effectiveness of our proposed sampling strategies against existing ones. So we mainly compare Stratified Sampling, Negative Sharing, and Stratified Sampling with Negative Sharing, against IID sampling and Negative Sampling. It is worth noting that several existing state-of-the-art models \cite{van2013deep,bansal2016ask,chen2017text} are special cases of our framework (e.g. using MSE-loss/Log-loss with CNN or RNN), so they are compared to other loss functions under our framework.

\paragraph{Evaluation Metrics} For recommendation performance, we follow \cite{wang2015collaborative,bansal2016ask} and use recall@M. As pointed out in \cite{wang2015collaborative}, the precision is not a suitable performance measure since non interactions may be due to (1) the user is not interested in the item, or (2) the user does not pay attention to its existence. More specifically, for each user, we rank candidate test items based on the predicted scores, and then compute recall@M based on the list. Finally the recall@M is averaged over all users. 

As for the computational cost, we mainly measure it in three dimensions: the training time for each iteration (or epoch equivalently, since batch size is fixed for all methods), the number of iterations needed to reach a referenced loss, and the total amount of computation time needed to reach the same loss. In our experiments, we use the smallest loss obtained by IID sampling in the maximum 30 epochs as referenced loss. Noted that all time measure mentioned here is in Wall Time.

\paragraph{Parameter Settings} The key parameters are tuned with validation set, while others are simply set to reasonable values. We adopt Adam \cite{kingma2014adam} as the stochastic optimizer. We use the same batch size $b=512$ for all sampling strategies, we use the number of positive link per sampled stratum $s=4$, learning rate is set to 0.001 for MSE-loss, and 0.01 for others. $\gamma$ is set to 0.1 for Hinge-loss, and 10 for others. $\lambda$ is set to 8 for MSE-loss, and 128 for others. We set number of negative examples $k=10$ for convolutional neural networks, and $k=5$ for RNN/LSTM due to the GPU memory limit. All experiments are run with Titan X GPUs. We use unigram noise/negative distribution.

For CNN, we adopt the structure similar in \cite{kim2014convolutional}, and use 50 filters with filter size of 3. Regularization is added using both weight decay on user embedding and dropout on item embedding. For RNN, we use LSTM \cite{hochreiter1997long} with 50 hidden units. For both models, the dimensions of user and word embedding are set to 50. Early stop is utilized, and the experiments are run to maximum 30 epochs.

\subsection{Speedup Under Different Sampling Strategies}

\begin{table}[t]
	\small
	\centering
	\caption{Comparisons of speedup for different sampling strategies against IID Sampling: per iteration, \# of iteration, and total speedup.}
	\label{tab:speedup}
	\begin{tabular}{|p{1.3em}c|ccc|ccc|}\hline
		&          & \multicolumn{3}{c|}{CiteULike}                                                                                              & \multicolumn{3}{c|}{News} \\ [5pt] 
		Model  & Sampling & \multicolumn{1}{c}{Per it.} & \multicolumn{1}{c}{\# of it.} & \multicolumn{1}{c|}{Total} & \multicolumn{1}{c}{Per it.} & \multicolumn{1}{c}{\# of it.} & \multicolumn{1}{c|}{Total}  \\[5pt] \hline
		\multirow{6}{*}{CNN}                & Negative    &            1.02 &           1.00 &            1.02  &            1.03 &           1.03 &            1.06\\[5pt]
		& Stratified&            8.83 &           0.97 &            8.56   &            6.40 &           0.97 &            6.20\\[5pt] 
		& N.S. &            8.42 &  \textbf{2.31} &           19.50 &            6.54 &  \textbf{2.21} &           14.45\\[5pt] 
		& Strat. w. N.S. &  \textbf{15.53} &           1.87 &  \textbf{29.12} &  \textbf{11.49} &           2.17 &  \textbf{24.98}\\[5pt] \hline
		\multirow{6}{*}{LSTM}                & Negative &             0.99 &             0.96 &           0.95&              1.0 &             1.25 &           1.25 \\[5pt]
		& Stratified  &              3.1 &             0.77 &           2.38 &             3.12 &             1.03 &           3.22\\[5pt] 
		& N.S. &             2.87 &    \textbf{2.45} &           7.03 &             2.78 &    \textbf{4.14} &  \textbf{11.5}\\[5pt] 
		& Strat. w. N.S. &     \textbf{3.4} &             2.22 &  \textbf{7.57}&    \textbf{3.13} &             3.32 &          10.41 \\[5pt] \hline
	\end{tabular}
\end{table}

\begin{figure*}[t]
	\centering
	\begin{subfigure}[b]{0.24\textwidth}
		\includegraphics[width=\textwidth]{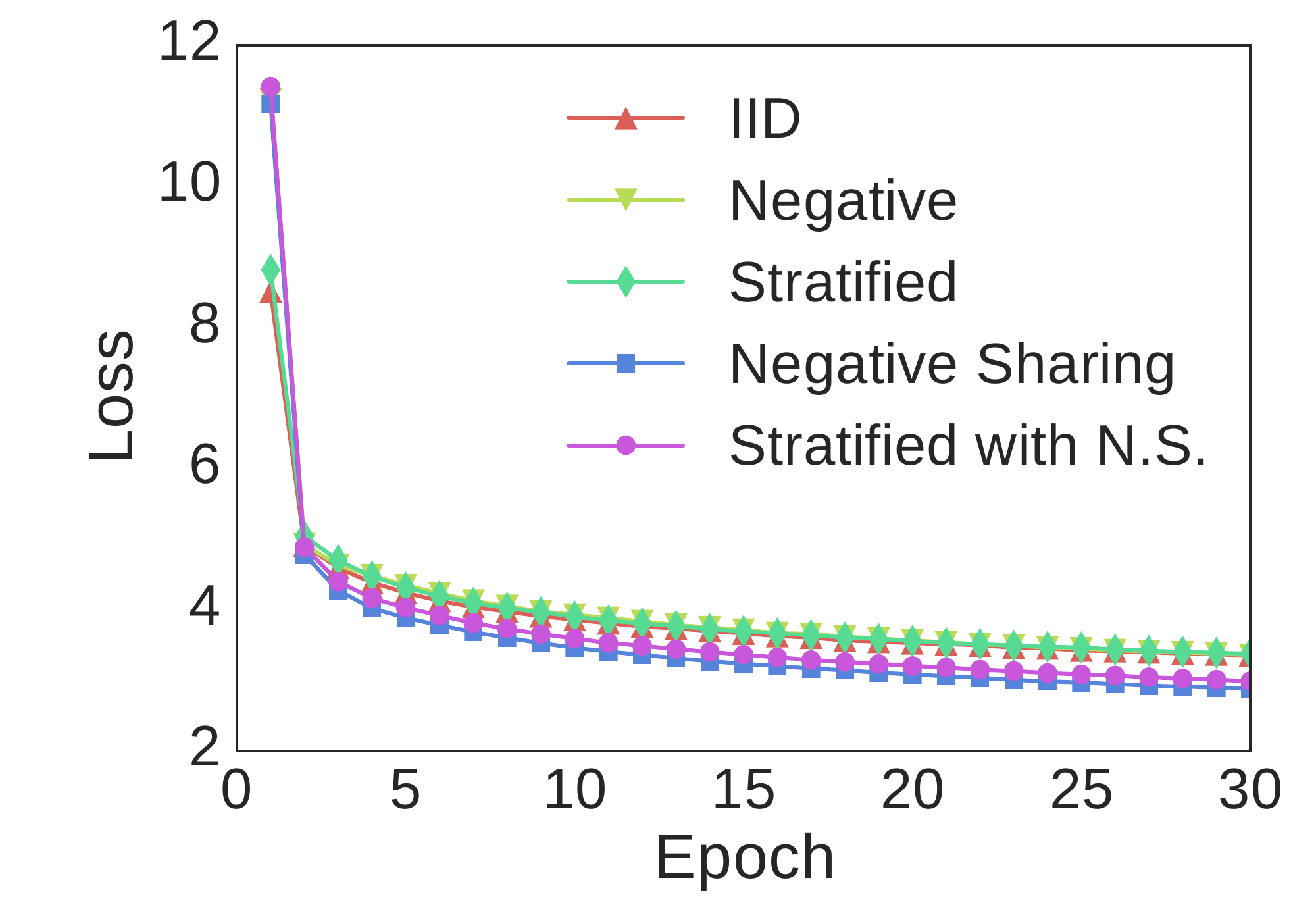}
		\caption{Citeulike (epoch)}
	\end{subfigure}
	\begin{subfigure}[b]{0.24\textwidth}
		\includegraphics[width=\textwidth]{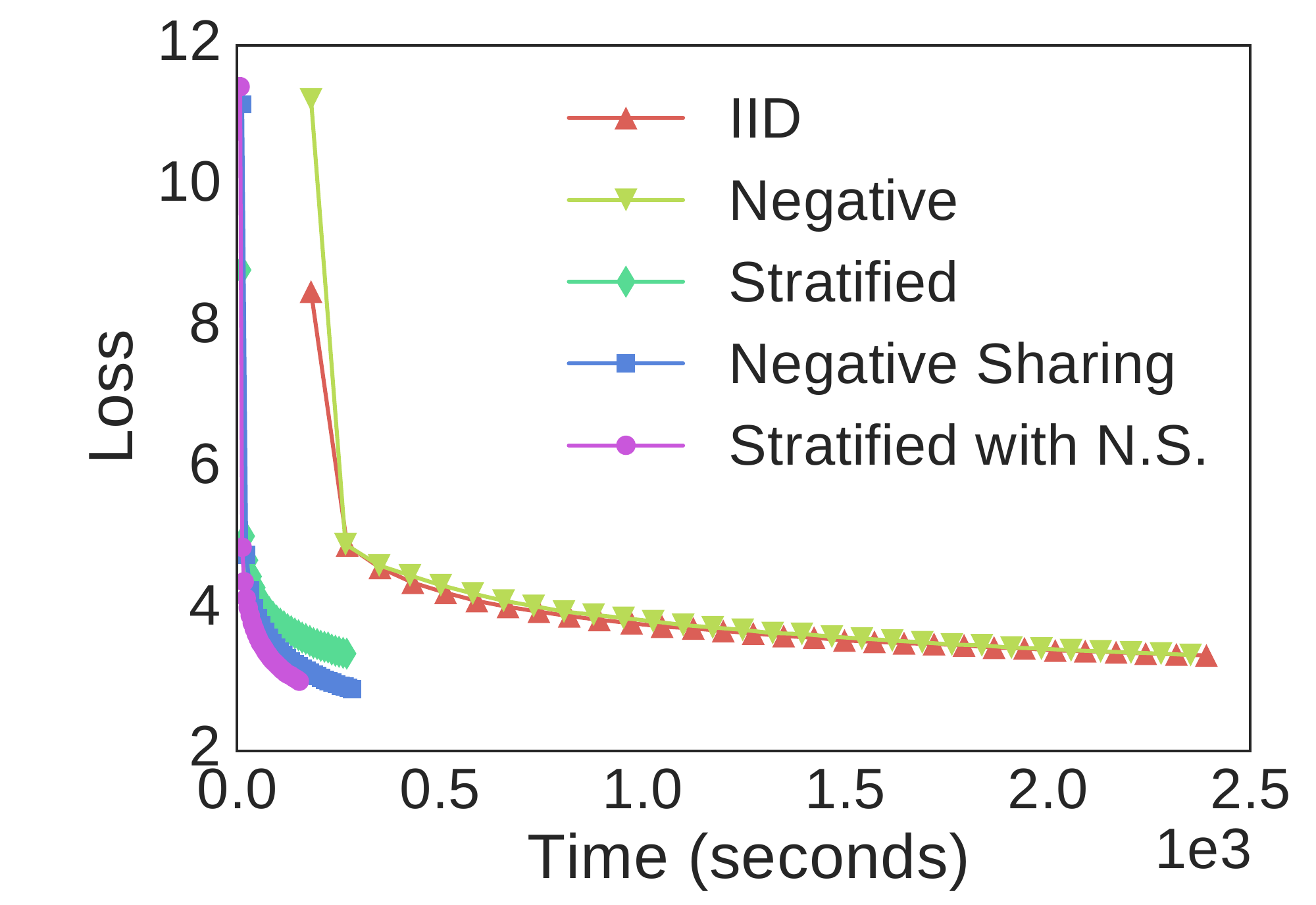}
		\caption{Citeulike (wall time)}
	\end{subfigure}
	\begin{subfigure}[b]{0.24\textwidth}
		\includegraphics[width=\textwidth]{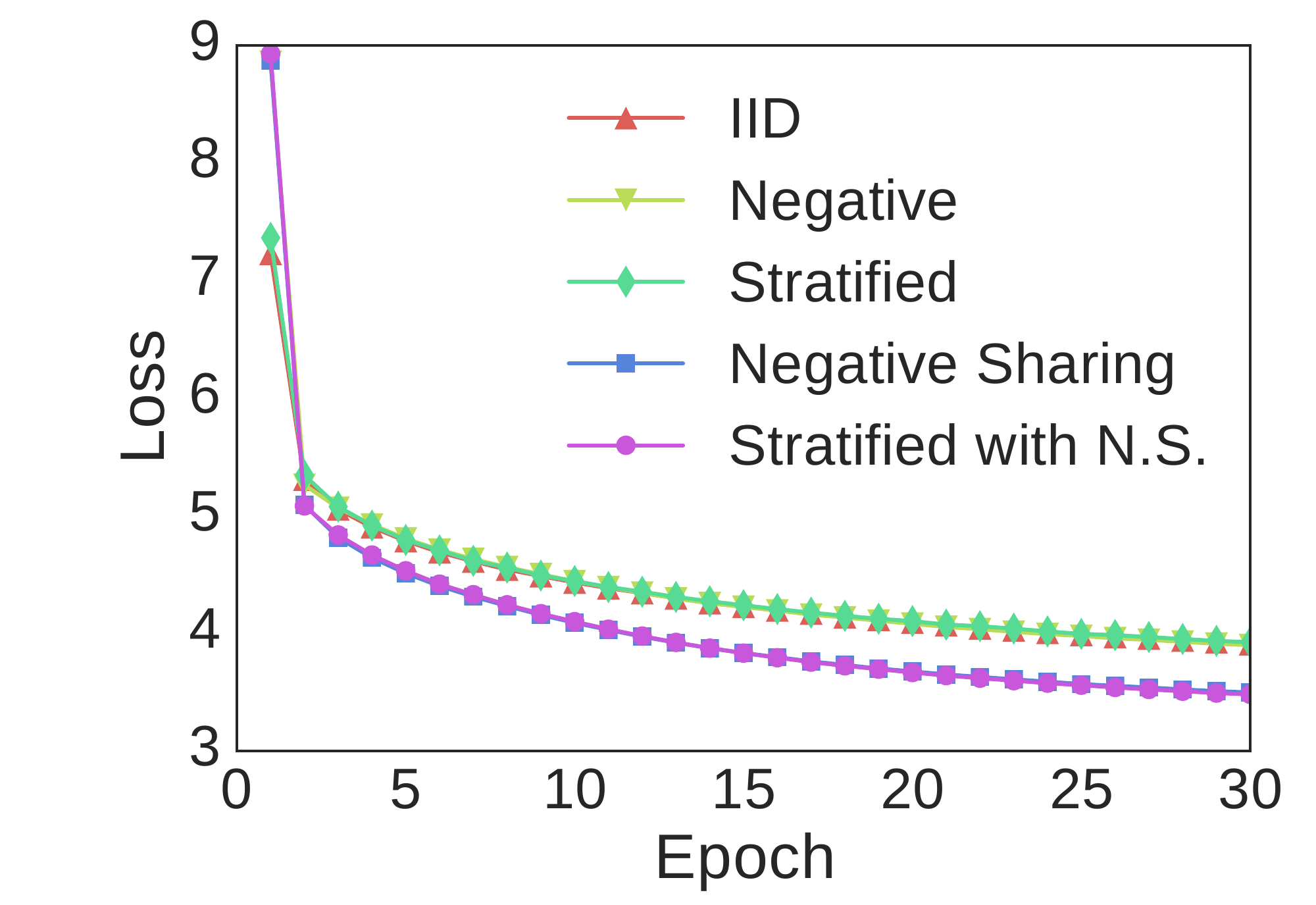}
		\caption{News (epoch)}
	\end{subfigure}
	\begin{subfigure}[b]{0.24\textwidth}
		\includegraphics[width=\textwidth]{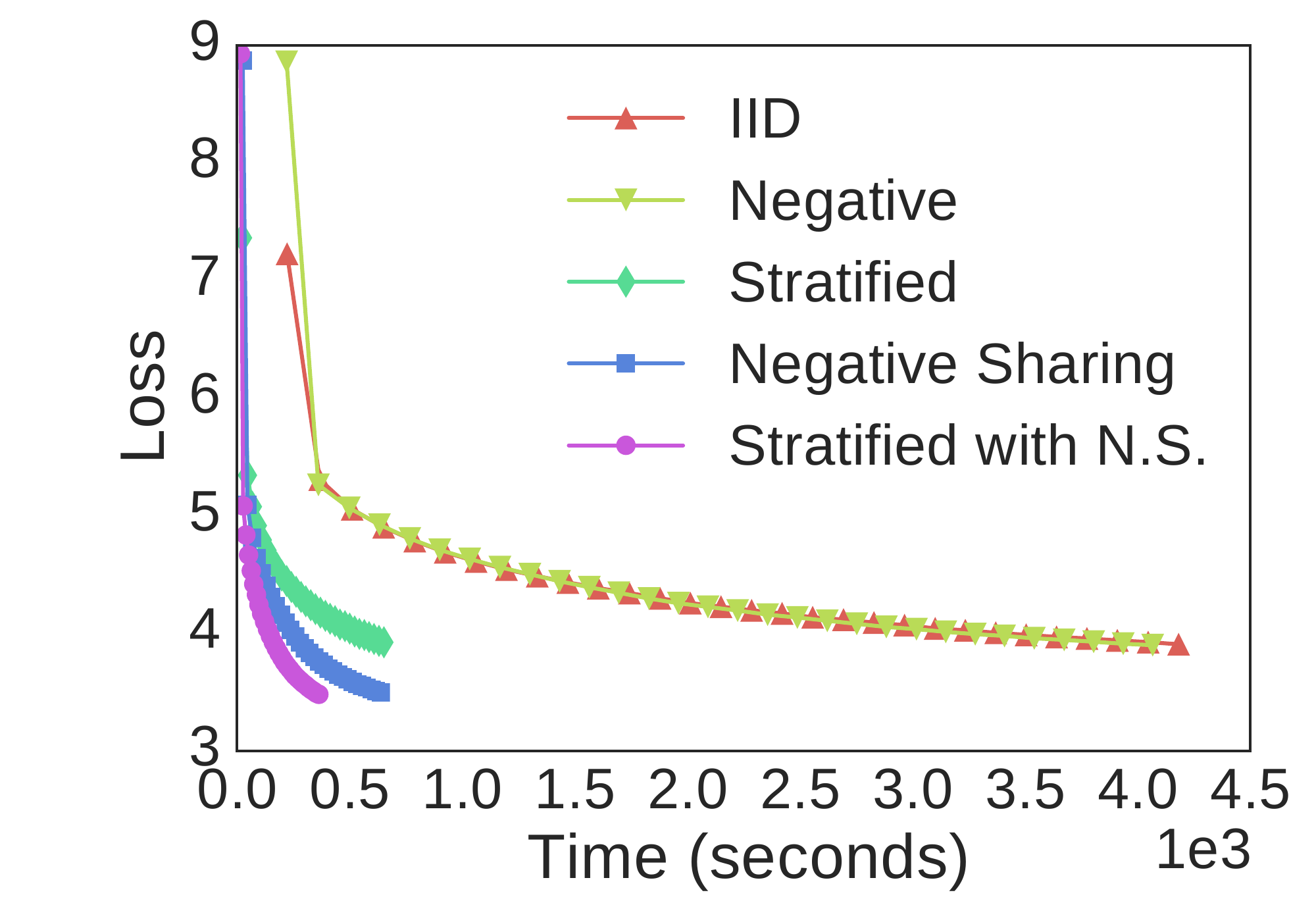}
		\caption{News (wall time)}
	\end{subfigure}
	\caption{Training loss curves (all methods have the same number of $b$ positive samples in a mini-batch)}
	\label{fig:convergence_curve_loss}
\end{figure*}
\begin{figure*}
	\begin{subfigure}[b]{0.24\textwidth}
		\includegraphics[width=\textwidth]{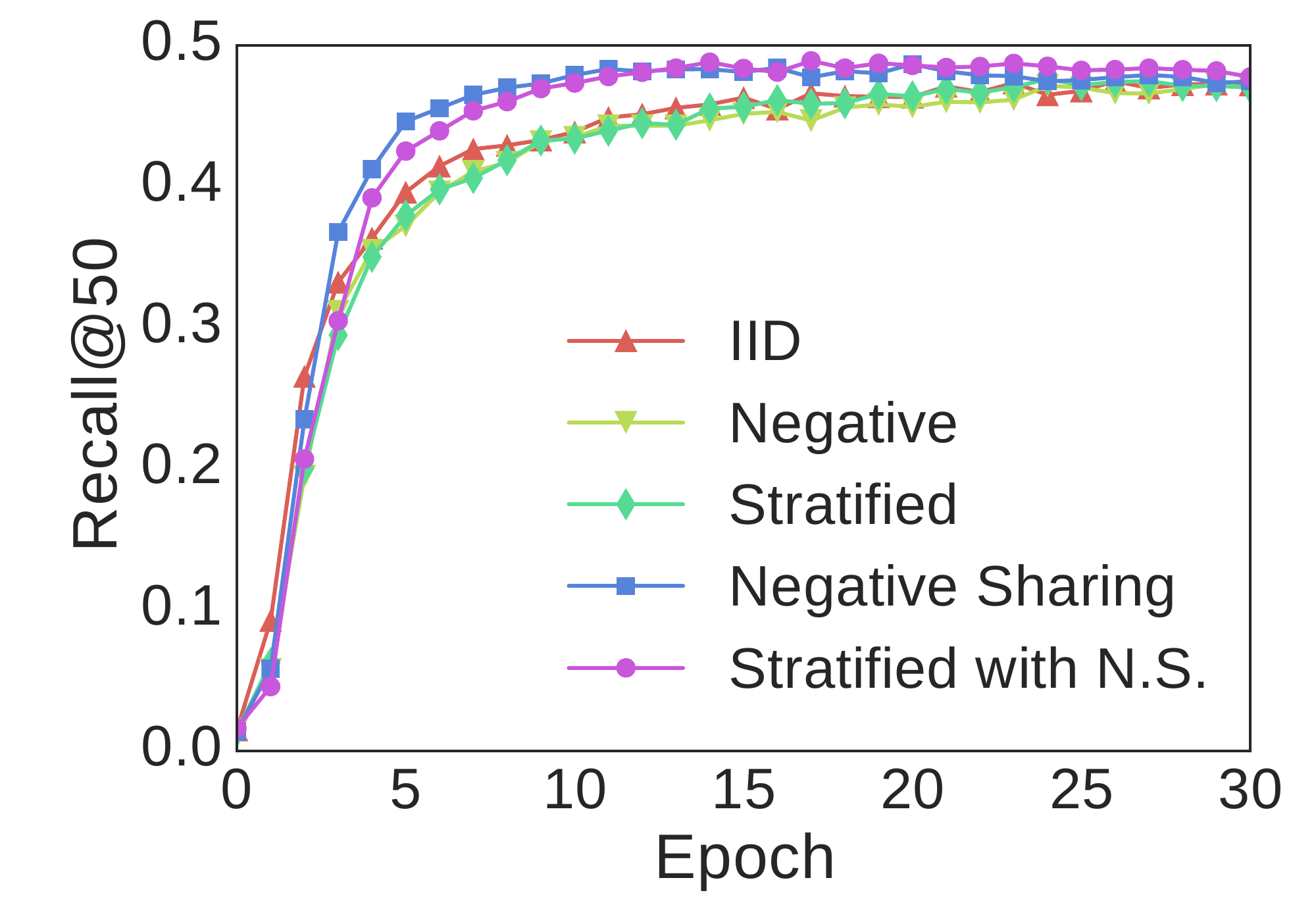}
		\caption{Citeulike (epoch)}
	\end{subfigure}
	\begin{subfigure}[b]{0.24\textwidth}
		\includegraphics[width=\textwidth]{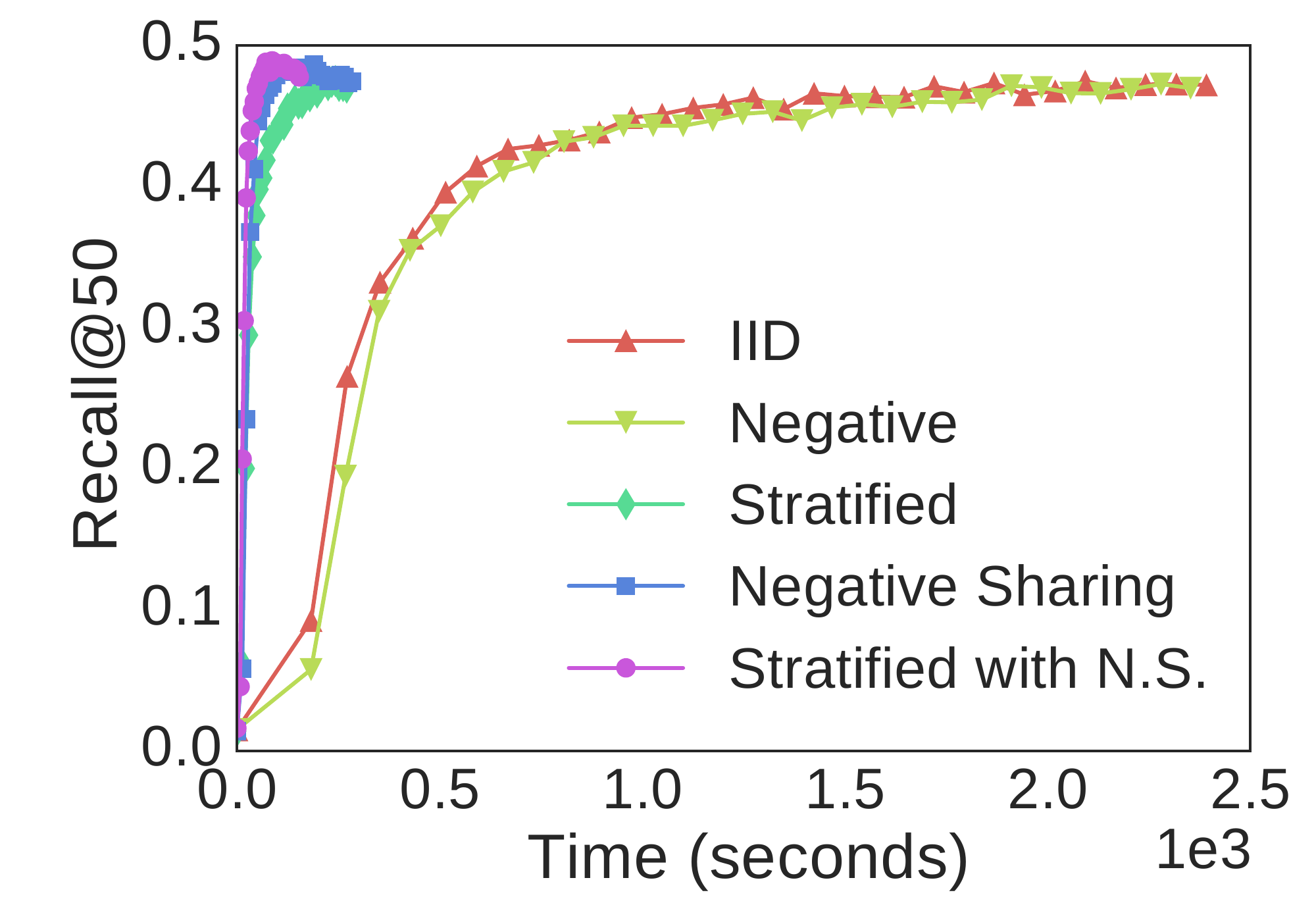}
		\caption{Citeulike (wall time)}
	\end{subfigure}
	\begin{subfigure}[b]{0.24\textwidth}
		\includegraphics[width=\textwidth]{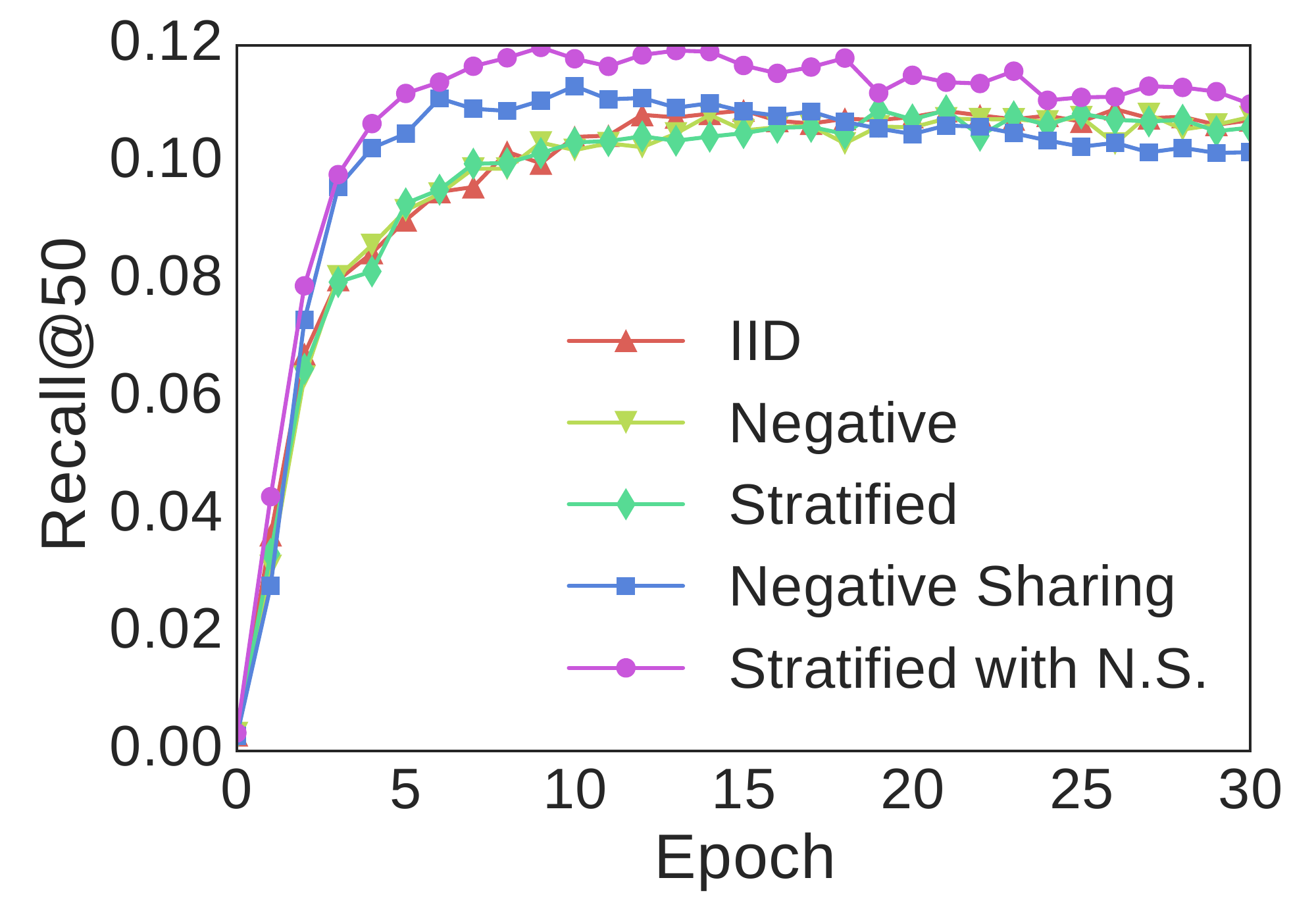}
		\caption{News (epoch)}
	\end{subfigure}
	\begin{subfigure}[b]{0.24\textwidth}
		\includegraphics[width=\textwidth]{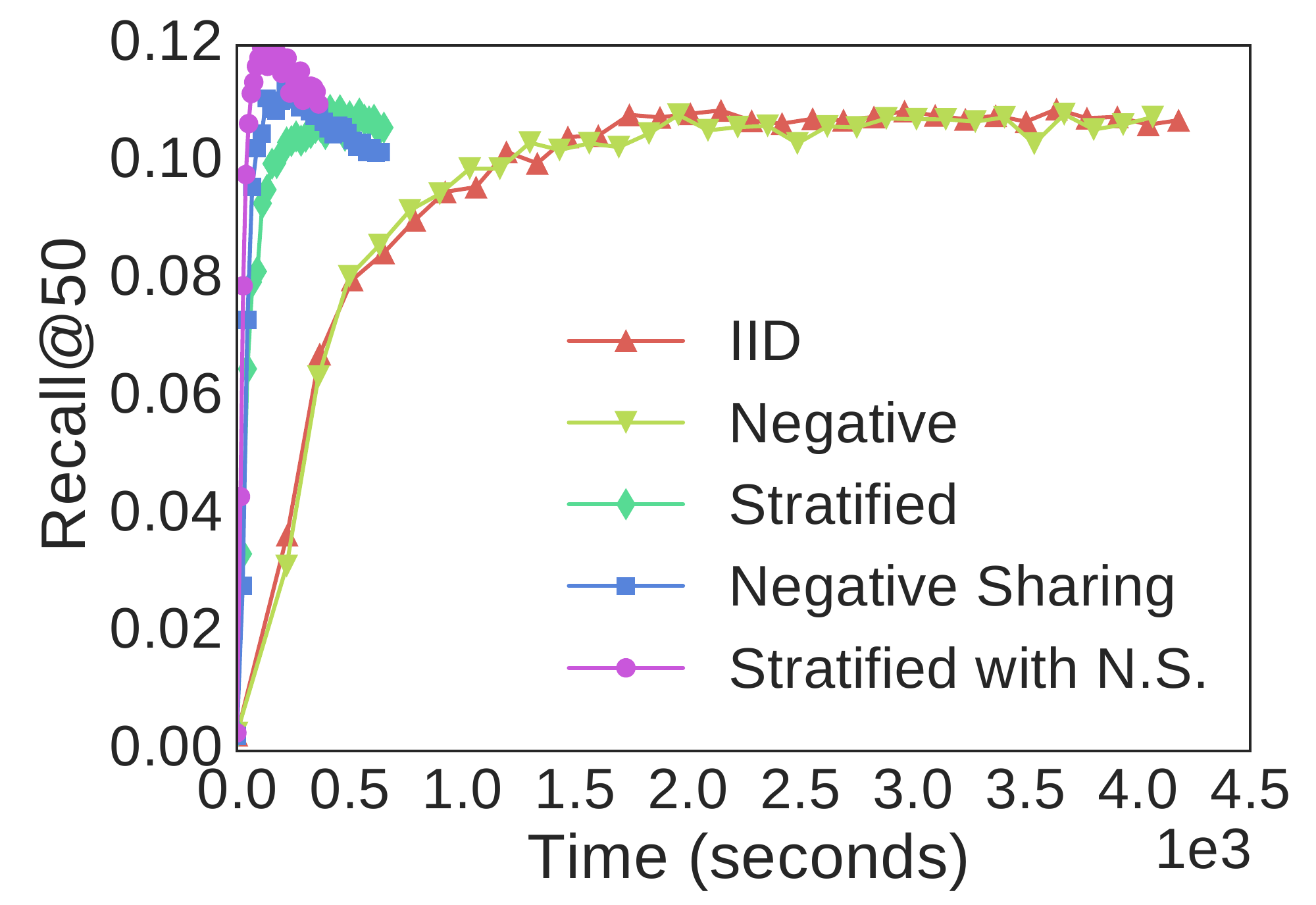}
		\caption{News (wall time)}
	\end{subfigure}
	\caption{Test performance/recall curves (all methods have the same number of $b$ positive samples in a mini-batch).}
	\label{fig:convergence_curve_recall}
\end{figure*}

Table \ref{tab:speedup} breaks down the speedup into (1) speedup for training on a given mini-batch, (2) number of iterations (to reach referenced cost) speedup, and (3) the total speedup, which is product of the first two. Different strategies are compared against IID Sampling. It is shown that Negative Sampling has similar computational cost as IID Sampling, which fits our projection. All three proposed sampling strategies can significantly reduce the computation cost within a mini-batch. Moreover, the Negative Sharing and Stratified Sampling with Negative Sharing can further improve the convergence w.r.t. the number of iterations, which demonstrates the benefit of using larger number of negative examples.

Figure \ref{fig:convergence_curve_loss} and \ref{fig:convergence_curve_recall} shows the convergence curves of both loss and test performance for different sampling strategies (with CNN + SG-loss). In both figures, we measure progress every epoch, which is equivalent to a fixed number of iterations since all methods have the same batch size $b$. In both figures, we can observe mainly two types of convergences behavior. Firstly, in terms of number of iterations, Negative Sharing (w./wo. Stratified Sampling) converge fastest, which attributes to the number of negative samples used. Secondly, in terms of wall time, Negative Sharing (w./wo. Stratified Sampling) and Stratified Sampling (by Items) are all significantly faster than baseline sampling strategies, i.e. IID Sampling and Neagtive Sampling. It is also interesting to see that that overfitting occurs earlier as convergence speeds up, which does no harm as early stopping can be used.

\begin{figure}[t!]
	\centering
	\begin{subfigure}[b]{0.23\textwidth}
		\includegraphics[width=\textwidth]{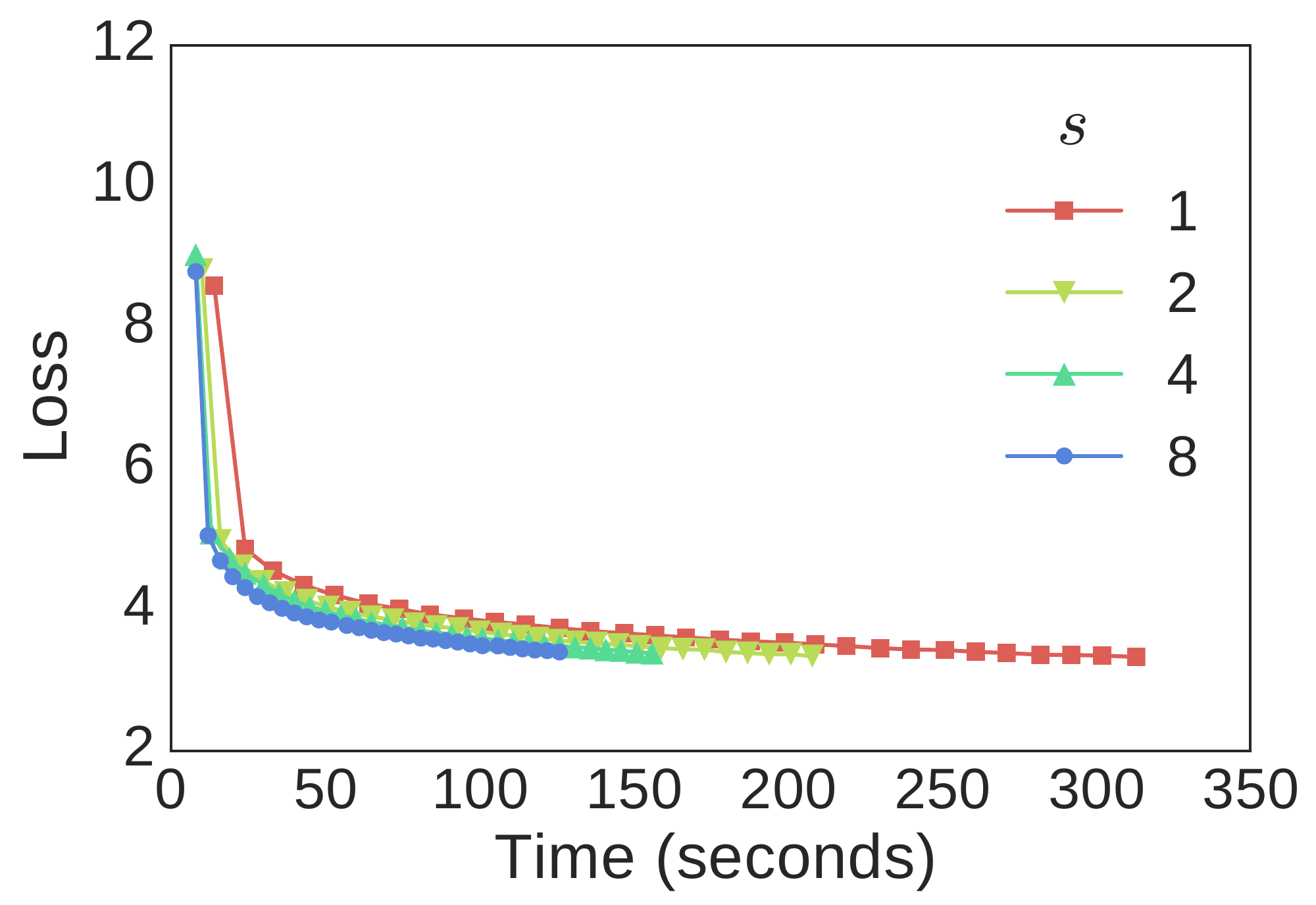}
		\caption{Loss (Stratified)}
		\label{}
	\end{subfigure}
	\begin{subfigure}[b]{0.23\textwidth}
		\includegraphics[width=\textwidth]{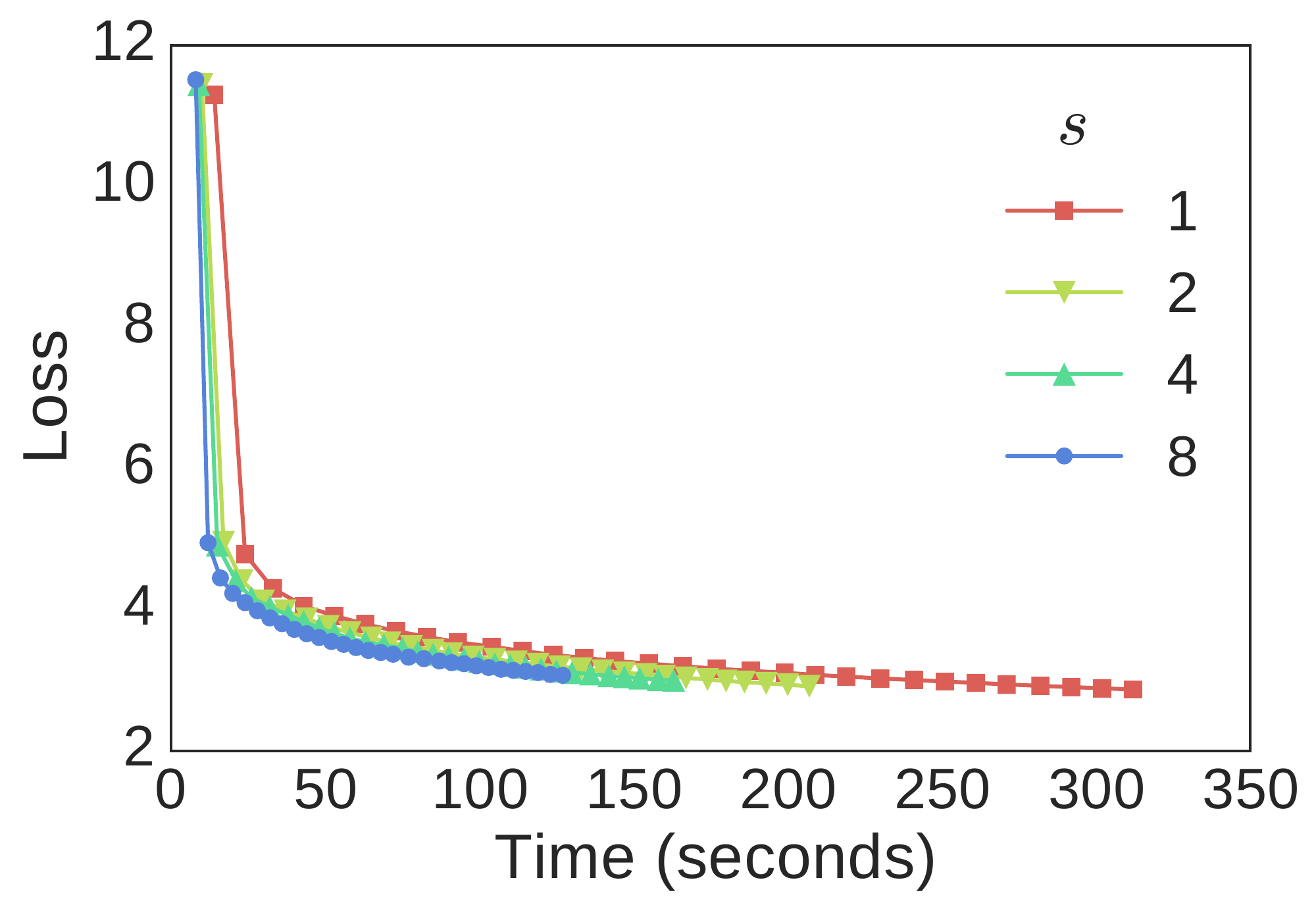}
		\caption{Loss (Stratified with N.S.)}
		\label{}
	\end{subfigure}
	\\
	\begin{subfigure}[b]{0.23\textwidth}
		\includegraphics[width=\textwidth]{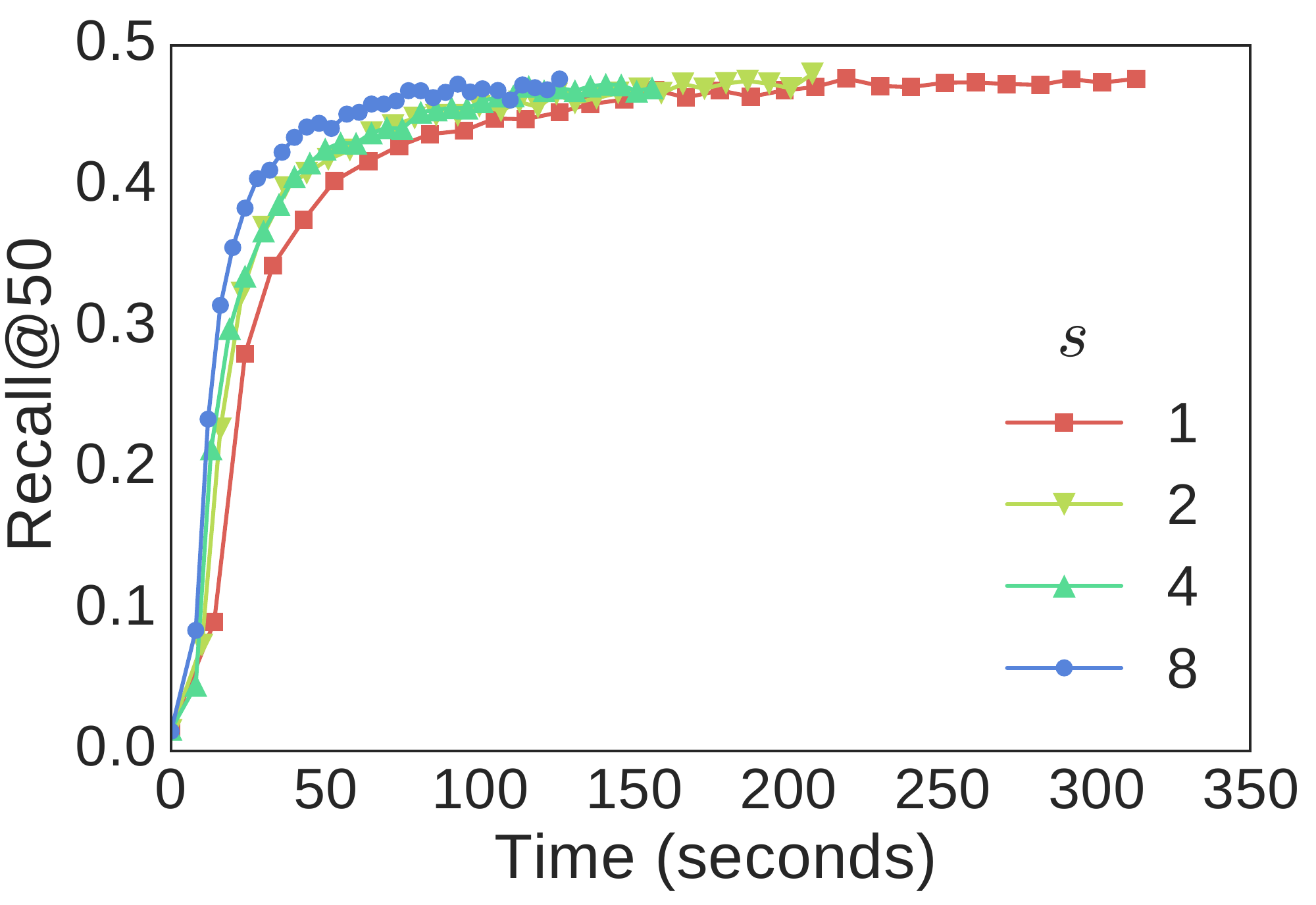}
		\caption{Recall (Stratified)}
		\label{}
	\end{subfigure}
	\begin{subfigure}[b]{0.23\textwidth}
		\includegraphics[width=\textwidth]{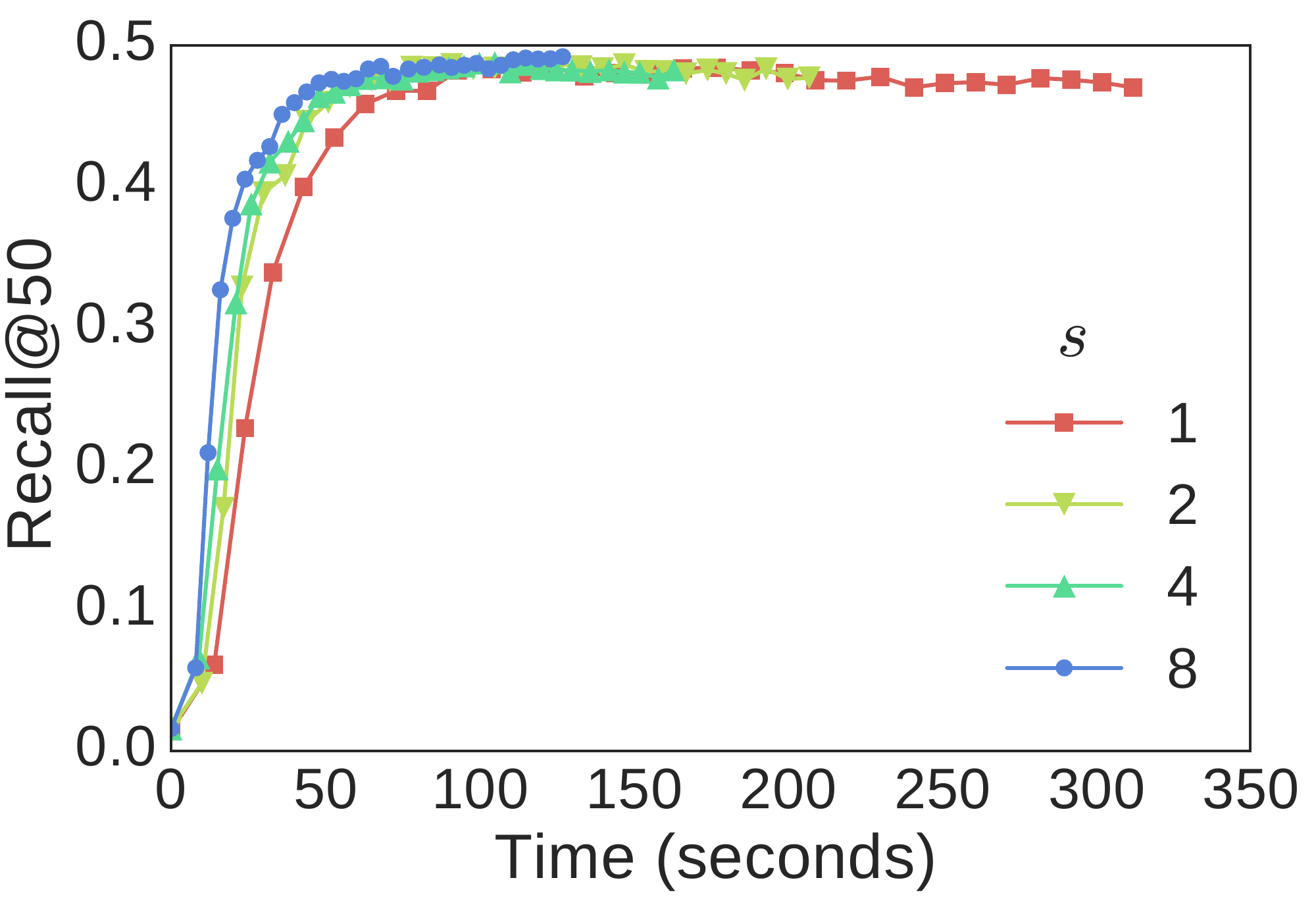}
		\caption{Recall (Stratified with N.S.)}
		\label{}
	\end{subfigure}
	\caption{The number of positive links per stratum $s$ VS loss and performance.}
	\label{fig:strata_size}
\end{figure}

For Stratified Sampling (w./wo. negative sharing), the number of positive links per stratum $s$ can also play a role to improve speedup as we analyzed before. As shown in Figure \ref{fig:strata_size}, the convergence time as well as recommendation performance can both be improved with a reasonable $s$, such as 4 or 8 in our case. 

\subsection{Recommendation Performance Under Different Sampling Strategies}

It is shown in above experiments that the proposed sampling strategies are significantly faster than the baselines. But we would also like to further access the recommendation performance by adopting the proposed strategies.

Table \ref{tab:perf_comp} compares the proposed sampling strategies with CNN/RNN models and four loss functions (both pointwise and pairwise). We can see that IID Sampling, Negative Sampling and Stratified Sampling (by Items) have similar recommendation performances, which is expected since they all utilize same amount of negative links. For Negative Sharing and Stratified Sampling with Negative Sharing, since there are much more negative samples utilized, their performances are significantly better. We also observe that the current recommendation models based on MSE-loss \cite{van2013deep,bansal2016ask} can be improved by others such as SG-loss and pairwise loss functions \cite{chen2017text}. 

\begin{table*}[t]
	\small
	\centering
	\caption{Recall@50 for different sampling strategies under different models and losses.}
	\label{tab:perf_comp}
	\begin{tabular}{|cc|cccc|cccc|}\hline
		 &          & \multicolumn{4}{c|}{CiteULike}                                                                                              & \multicolumn{4}{c|}{News} \\ [5pt] 
		Model  & Sampling & \multicolumn{1}{l}{SG-loss} & \multicolumn{1}{l}{MSE-loss} & \multicolumn{1}{l}{Hinge-loss} & \multicolumn{1}{l|}{Log-loss}& \multicolumn{1}{l}{SG-loss} & \multicolumn{1}{l}{MSE-loss} & \multicolumn{1}{l}{Hinge-loss} & \multicolumn{1}{l|}{Log-loss} \\[5pt] \hline
		\multirow{7}{*}{CNN}                & IID &           0.4746 &           0.4437 &           - &           - &           0.1091 &           0.0929 &           - &           -  \\[5pt] 
		& Negative        &           0.4725 &           0.4408 &           0.4729 &           0.4796  &           0.1083 &           0.0956 &           0.1013 &           0.1009\\[5pt]
		& Stratified &           0.4761 &           0.4394 &           - &           - &           0.1090 &           0.0913 &           - &           - \\[5pt] 
		& Negative Sharing &           0.4866 &           0.4423 &  \textbf{0.4794} &           0.4769&           0.1131 &           0.0968 &           0.0909 &           0.0932\\[5pt] 
		& Stratified with N.S. & \textbf{0.4890} &  \textbf{0.4535} &           0.4790 &  \textbf{0.4884}&  \textbf{0.1196} &  \textbf{0.1043} &  \textbf{0.1059} &  \textbf{0.1100} \\[5pt] \hline
		\multirow{7}{*}{LSTM}                
		& IID  &           0.4479 &           0.4718 &           - &           - &           0.0971 &           0.0998 &           - &           - \\[5pt] 
		& Negative &           0.4371 &           0.4668 &           0.4321 &           0.4540  &           0.0977 &           0.0977 &           0.0718 &           0.0711\\[5pt]
		& Stratified  &           0.4344 &           0.4685 &           - &           -  &           0.0966 &           0.0996 &           - &           - \\[5pt] 
		& Negative Sharing &           0.4629 &           0.4839 &           0.4605 &           0.4674  &  \textbf{0.1121} &           0.0982 &           0.0806 &           0.0862  \\[5pt] 
		& Stratified with N.S. &  \textbf{0.4742} &  \textbf{0.4877} &  \textbf{0.4703} &  \textbf{0.4730} &           0.1051 &  \textbf{0.1098} &  \textbf{0.1017} &  \textbf{0.1002}\\[5pt] \hline
	\end{tabular}
\end{table*}

\begin{figure}[t!]
	\centering
	\begin{subfigure}[b]{0.23\textwidth}
		\includegraphics[width=\textwidth]{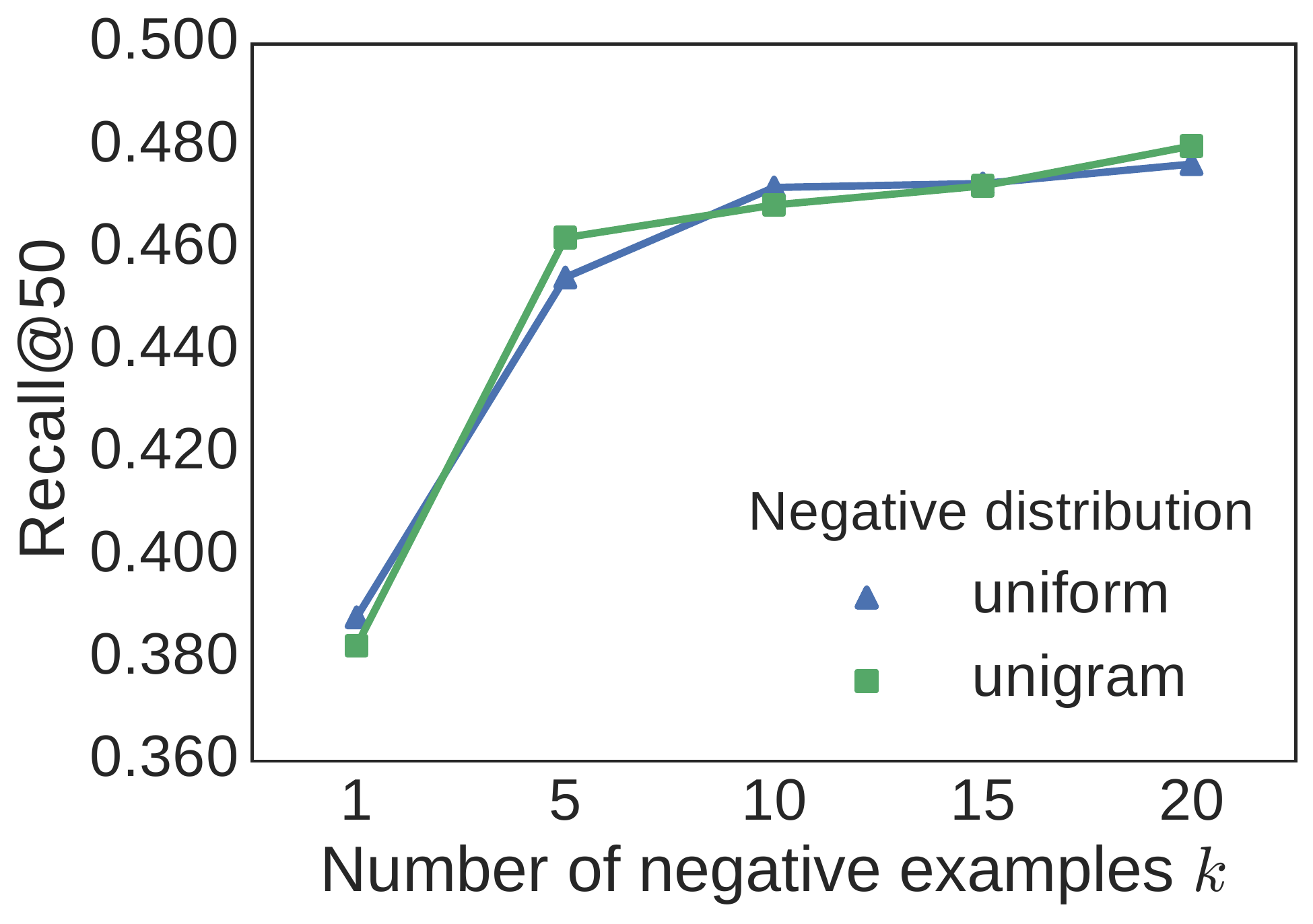}
		\caption{CiteULike}
		\label{}
	\end{subfigure}
	\begin{subfigure}[b]{0.23\textwidth}
		\includegraphics[width=\textwidth]{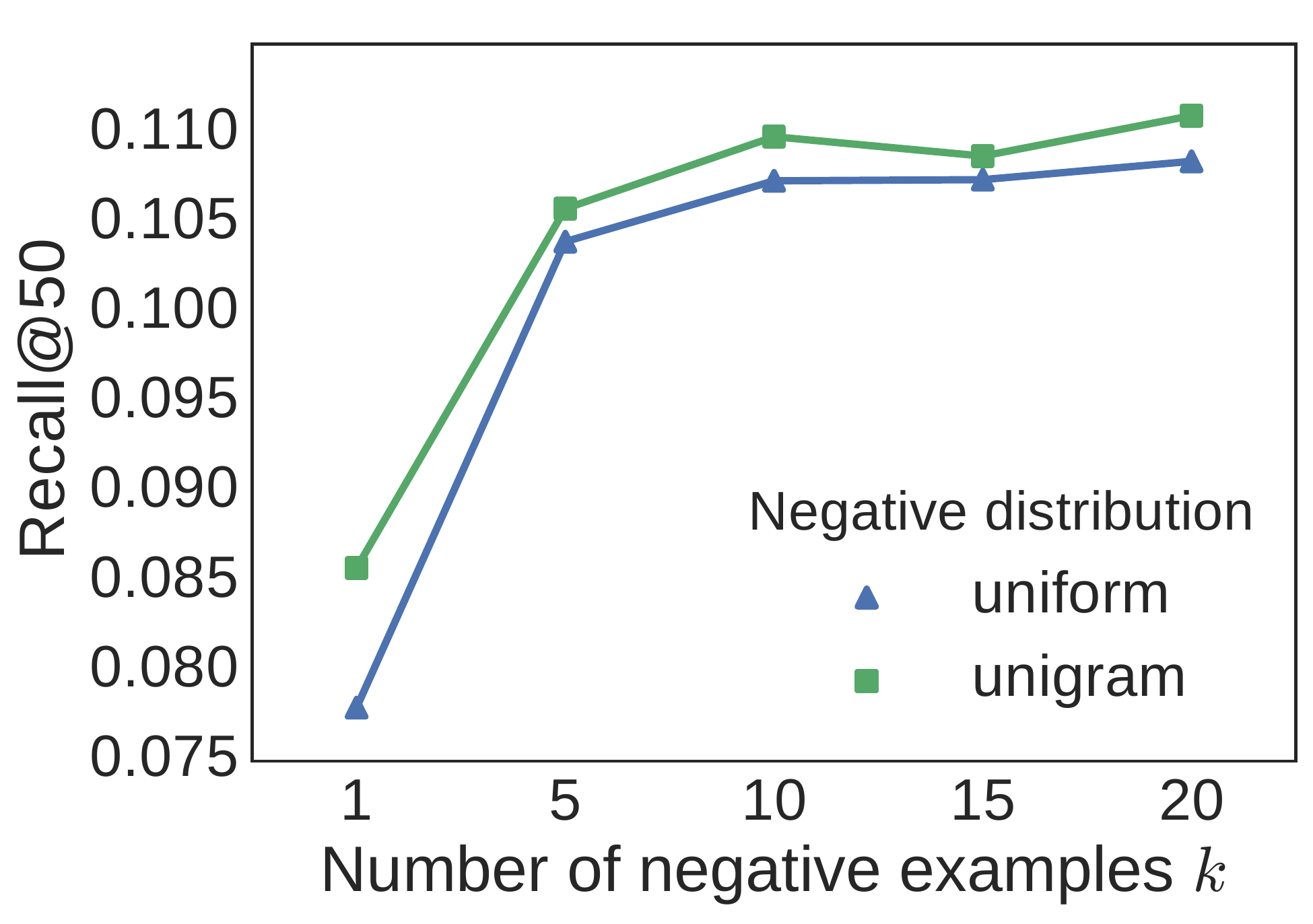}
		\caption{News}
		\label{}
	\end{subfigure}
	\caption{The number of negatives VS performances.}
	\label{fig:negnums}
\end{figure}

To further investigate the superior performance brought by Negative Sharing. We study the number of negative examples $k$ and the convergence performance. Figure \ref{fig:negnums} shows the test performance against various $k$. As shown in the figure, we observe a clear diminishing return in the improvement of performance. However, the performance seems still increasing even we use 20 negative examples, which explains why our proposed method with negative sharing can result in better performance.

\section{Related Work}

Collaborative filtering \cite{koren2009matrix} has been one of the most effective methods in recommender systems, and methods like matrix factorization \cite{koren2008factorization,salakhutdinov2011probabilistic} are widely adopted. While many papers focus on the explicit feedback setting such as rating prediction, implicit feedback is found in many real-world scenarios and studied by many papers as well \cite{pan2008one,hu2008collaborative,rendle2009bpr}. Although collaborative filtering techniques are powerful, they suffer from the so-called ``cold-start'' problem since side/content information is not well leveraged. To address the issue and improve performance, hybrid methods are proposed to incorporate side information \cite{singh2008relational,rendle2010factorization,zhou2011functional,chen2012svdfeature,chen2017task}, as well as content information \cite{wang2011collaborative,gopalan2014content,wang2015collaborative,chen2017text}.

Deep Neural Networks (DNNs) have been showing extraordinary abilities to extract high-level features from raw data, such as video, audio, and text \cite{collobert2011natural,kim2014convolutional,zhang2015character}. Compared to traditional feature detectors, such as SIFT and n-grams, DNNs and other embedding methods \cite{tang2015line,chen2016entity,chen2017task} can automatically extract better features that produce higher performance in various tasks. To leverage the extraordinary feature extraction or content understanding abilities of DNNs for recommender systems, recent efforts are made in combining collaborative filtering and neural networks \cite{van2013deep,wang2015collaborative,bansal2016ask,chen2017text}. \cite{wang2015collaborative} adopts autoencoder for extracting item-side text information for article recommendation, \cite{bansal2016ask} adopts RNN/GRU to better understand the text content. \cite{chen2017text} proposes to use CNN and pairwise loss functions, and also incorporate unsupervised text embedding. The general functional embedding framework in this work subsumes existing models \cite{van2013deep,bansal2016ask,chen2017text}.

Stochastic Gradient Descent \cite{bottou2010large} and its variants \cite{kingma2014adam} have been widely adopted in training machine learning models, including neural networks. Samples are drawn uniformly at random (IID) so that the stochastic gradient vector equals to the true gradient in expectation. In the setting where negative examples are overwhelming, such as in word embedding (e.g., Word2Vec \cite{mikolov2013distributed}) and network embedding (e.g., LINE \cite{tang2015line}) tasks, negative sampling is utilized. Recent efforts have been made to improve SGD convergence by (1) reducing the variance of stochastic gradient estimator, or (2) distributing the training over multiple workers. Several sampling techniques, such as stratified sampling \cite{zhao2014accelerating} and importance sampling \cite{zhao2015stochastic} are proposed to achieve the variance reduction. Different from their work, we improve sampling strategies in SGD by reducing the computational cost of a mini-batch while preserving, or even increasing, the number of data points in the mini-batch. Sampling techniques are also studied in \cite{gemulla2011large,zhuang2013fast} to distribute the computation of matrix factorization, their objectives in sampling strategy design are reducing the parameter overlapping and cache miss. We also find that the idea of sharing negative examples is exploited to speed up word embedding training in \cite{ji2016parallelizing}.

\section{Discussions}

While it is discussed under content-based collaborative filtering problem in this work, the study of sampling strategies for ``graph-based'' loss functions have further implications. The IID sampling strategy is simple and popular for SGD-based training, since the loss function terms usually do not share the common computations. So no matter how a mini-batch is formed, it almost bears the same amount of computation. This assumption is shattered by models that are defined under graph structure, with applications in social and knowledge graph mining \cite{bordes2013translating}, image caption ranking \cite{lin2016leveraging}, and so on. For those scenarios, we believe better sampling strategies can result in much faster training than that with IID sampling.

We would also like to point out limitations of our work. The first one is the setting of implicit feedback. When the problem is posed under explicit feedback, Negative Sharing can be less effective since the constructed negative samples may not overlap with the explicit negative ones. The second one is the assumption of efficient computation for interaction functions. When we use neural networks as interaction functions, we may need to consider constructing negative samples more wisely for Negative Sharing as it will also come with a noticeable cost.

\section{Conclusions and Future Work}

In this work, we propose a hybrid recommendation framework, combining conventional collaborative filtering with (deep) neural networks. The framework generalizes several existing state-of-the-art recommendation models, and embody potentially more powerful ones. To overcome the high computational cost brought by combining ``cheap'' CF with ``expensive'' NN, we first establish the connection between the loss functions and the user-item interaction bipartite graph, and then point out the computational costs can vary with different sampling strategies. Based on this insight, we propose three novel sampling strategies that can significantly improve the training efficiency of the proposed framework, as well as the recommendation performance.

In the future, there are some promising directions. Firstly, based on the efficient sampling techniques of this paper, we can more efficiently study different neural networks and auxiliary information for building hybrid recommendation models. Secondly, we can also study the effects of negative sampling distributions and its affect on the design of more efficient sampling strategies. Lastly but not least, it would also be interesting to apply our sampling strategies in a distributed training environments where multi-GPUs and multi-machines are considered.

\section*{Acknowledgements}

The authors would like to thank anonymous reviewers for helpful suggestions. The authors would also like to thank NVIDIA for the donation of one Titan X GPU. This work is partially supported by NSF CAREER \#1741634.

\bibliographystyle{ACM-Reference-Format}
\bibliography{main}

\newpage
\clearpage
\onecolumn
\appendix
\section*{Supplementary Material}
\resetcounters 

\section{Proofs}

Here we give the proofs for both the lemma and the proposition introduced in the main paper. For brevity, throughout we assume by default the loss function $\mathcal{L}$ is the pointwise loss of Eq. (1) in the main paper. Proofs are only given for the pointwise loss, but it can be similarly derived for the pairwise loss. We start by first introducing some definitions.
\begin{definition0}
	A function $f$ is $L$-$smooth$ if there is a constant $L$ such that
	$$ \|\nabla f(x) - \nabla f(y)\| \le L \|x - y\|
	$$
\end{definition0}
Such an assumption is very common in the analysis of first-order methods. In the following proof, we assume any loss functions $\mathcal{L}$ is $L$-$smooth$.
\begin{property0}
	(Quadratic Upper Bound) A $L$-$smooth$ function $f$ has the following property
	$$
	f(y) \le f(x) + \nabla f(x)^T (y - x) + \frac{L}{2} \|y - x\|^2
	$$
\end{property0}
\begin{definition0}
	We say a function $f$ has a $\sigma$-$bounded$ gradient if $\|\nabla f_i(\theta)\|^2 \le \sigma$ for all $i\in [n]$ and any $\theta \in \mathbb{R}^d$.
\end{definition0}
For each training iteration, we first sample a mini-batch of links (denoted by $B$) of both positive links ($B^+$) and negative links ($B^-$), according to the sampling algorithm (one of the Algorithm 2, 3, 4, 5), and then the stochastic gradient is computed and applied to the parameters as follows:
\begin{equation}
\label{eq:update}
\mathbf{\theta}^{t + 1} = \mathbf{\theta}^t  - \frac{\eta_t}{m} \sum_{(u, v) \in B^+_t} c^+_{uv}\nabla \mathcal{L}^+(\theta|u, v) - \frac{\eta_t}{n} \sum_{(u, v) \in B^-_t} c^-_{uv}\nabla \mathcal{L}^-(\theta|u, v)
\end{equation}
Here we use $\mathcal{L}^+(\theta|u, v)$ to denote the gradient of loss function $\mathcal{L}^+(\theta)$ given a pair of $(u, v)$. And $m$, $n$ are the number of positive and negative links in the batch $B$, respectively.
\begin{lemma0}
	\label{th:convergence_lemma} (unbiased stochastic gradient)
	Under sampling Algorithm 2, 3, 4, 5, we have $\mathbb{E}_B[\nabla \mathcal{L}_{B}(\theta^t)] = \nabla\mathcal{L}(\theta^t)$. In other words, the stochastic mini-batch gradient equals to true gradient in expectation.
\end{lemma0}
\begin{proof}
	Below we prove this lemma for each for the sampling Algorithm. For completeness, we also show the proof for Uniform Sampling as follows. The main idea is show the expectation of stochastic gradient computed in a randomly formed mini-batch equal to the true gradient of objective in Eq. \ref{eq:point}.
	
	\paragraph{IID Sampling}
	The positive links in the batch $B$ are i.i.d. samples from $P_d(u,v)$ (i.e. drawn uniformly at random from all positive links), and the negative links in $B$ are i.i.d. samples from $P_d(u) P_n(v)$, thus we have
	\begin{equation}
	\begin{split}
	 & \mathbb{E}_B[\nabla \mathcal{L}_{B}(\theta^t)] \\
	=& \frac{1}{m}\sum_{i=1}^{m}\mathbb{E}_{(u, v)\sim P_d(u, v)}[c^+_{uv}\nabla\mathcal{L}^+(\theta|u, v)] + \frac{1}{n}\sum_{i=1}^{n}\mathbb{E}_{(u, v)\sim P_d(u) P_n(v')}[c^-_{uv'} \nabla\mathcal{L}^-(\theta|u, v')]\\
	=& \mathbb{E}_{u\sim P_d(u)}\bigg[\mathbb{E}_{v\sim P_d(v|u)}[ c^{+}_{uv}\nabla\mathcal{L}^+(\theta|u, v)] + \mathbb{E}_{v'\sim P_n(v')}[c^{-}_{uv'} \nabla\mathcal{L}^-(\theta|u, v')]\bigg]\\
	=&\nabla \mathcal{L}(\theta^t)
	\end{split}
	\end{equation}
	The first equality is due to the definition of sampling procedure, the second equality is due to the definition of expectation, and the final equality is due to the definition of pointwise loss function in Eq. \ref{eq:point}.
	
	\paragraph{Negative Sampling}
	In Negative Sampling, we have batch $B$ consists of i.i.d. samples of $m$ positive links, and conditioning on each positive link, $k$ negative links are sampled by replacing items in the same i.i.d. manner. Positive links are sampled from $P_d(u, v)$, and negative items are sampled from $P_n(v')$, thus we have
	\begin{equation}
	\begin{split}
	& \mathbb{E}_B[\nabla \mathcal{L}_{B}(\theta^t)] \\
	=& \frac{1}{m}\sum_{i=1}^{m}\mathbb{E}_{(u, v)\sim P_d(u, v)} \frac{1}{k}\sum_{j=1}^{k} \mathbb{E}_{v'\sim P_n(v')}[c^+_{uv}\nabla\mathcal{L}^+(\theta|u, v) + c^-_{uv'} \nabla\mathcal{L}^-(\theta|u, v')]\\
	=& \mathbb{E}_{u\sim P_d(u)}\bigg[
	\mathbb{E}_{v\sim P_d(v|u)}[c^{+}_{uv}\nabla\mathcal{L}^+(\theta|u, v)] + \mathbb{E}_{v'\sim P_n(v')}[c^{-}_{uv} \nabla\mathcal{L}^-(\theta|u, v')] \bigg]\\
	=& \nabla \mathcal{L}(\theta^t)
	\end{split}
	\end{equation}
	The first equality is due to the definition of sampling procedure, and the second equality is due to the properties of joint probability distribution and expectation.
	
	\paragraph{Stratified Sampling (by Items)}
	In Stratified Sampling (by Items), a batch $B$ consists of links samples drawn in two steps: (1) draw an item $v\sim P_d(v)$, and (2) draw positive users $u\sim P_d(u|v)$ and negative users $u'\sim P_d(u)$ respectively. Additionally, negative terms are also re-weighted, thus we have
	\begin{equation}
	\begin{split}
	& \mathbb{E}_B[\nabla \mathcal{L}_{B}(\theta^t)] \\
	=& \mathbb{E}_{v\sim P_d(v)}\bigg[
	\frac{1}{m} \sum_{i=1}^{m}\mathbb{E}_{u\sim P_d(u|v)}[c^{+}_{uv}\nabla\mathcal{L}^+(\theta|u, v)] + \frac{1}{n}\sum_{i=1}^{n}\mathbb{E}_{u\sim P_d(u)}[c^{-}_{uv} \frac{P_n(v)}{P_d(v)} \nabla\mathcal{L}^-(\theta|u, v)] \bigg]\\
	=& \mathbb{E}_{(u, v)\sim P_d(u, v)} [c^{+}_{uv}\nabla\mathcal{L}^+(\theta|u, v)] + \mathbb{E}_{(u, v)\sim P_d(u) P_d(v)}[c^{-}_{uv} \frac{P_n(v)}{P_d(v)} \nabla\mathcal{L}^-(\theta|u, v)] \\
	=& \mathbb{E}_{(u, v)\sim P_d(u, v)} [c^{+}_{uv}\nabla\mathcal{L}^+(\theta|u, v)] + \mathbb{E}_{(u, v)\sim P_d(u) P_n(v)}[c^{-}_{uv} \nabla\mathcal{L}^-(\theta|u, v)] \\
	=& \mathbb{E}_{u\sim P_d(u)}\bigg[
	\mathbb{E}_{v\sim P_d(v|u)}[c^{+}_{uv}\nabla\mathcal{L}^+(\theta|u, v)] + \mathbb{E}_{v'\sim P_n(v')}[c^{-}_{uv} \nabla\mathcal{L}^-(\theta|u, v')] \bigg]\\
	=& \nabla \mathcal{L}(\theta^t)
	\end{split}
	\end{equation}
	The first equality is due to the definition of sampling procedure, and the second, the third and the forth equality is due to the properties of joint probability distribution and expectation.
	
	\paragraph{Negative Sharing}
	In Negative Sharing, we only draw positive links uniformly at random (i.e. $(u,v)\sim P_d(u, v)$), while constructing negative links from sharing the items in the batch. So the batch $B$ we use for computing gradient consists of both $m$ positive links and $m(m-1)$ negative links.
	
	Although we do not draw negative links directly, we can still calculate their probability according to the probability distribution from which we draw the positive links. So a pair of constructed negative link in the batch is drawn from $(u, v)\sim P_d(u, v) = P_d(v) P_d(u|v)$. Additionally, negative terms are also re-weighted, we have
	\begin{equation}
	\begin{split}
	& \mathbb{E}_B[\nabla \mathcal{L}_{B}(\theta^t)] \\
	=& \frac{1}{m}\sum_{i=1}^{m}\mathbb{E}_{(u, v)\sim P_d(u, v)} [c^+_{uv}\nabla\mathcal{L}^+(\theta|u, v)] + \frac{1}{m(m-1)}\sum_{j=1}^{m(m-1)}\mathbb{E}_{(u, v)\sim P_d(u, v)} [c^-_{uv} \frac{P_n(v)}{P_d(v)}\nabla\mathcal{L}^-(\theta|u, v)]\\
	=& \mathbb{E}_{u\sim P_d(u)}\bigg[
	\mathbb{E}_{v\sim P_d(v|u)}[c^{+}_{uv}\nabla\mathcal{L}^+(\theta|u, v)] + \mathbb{E}_{v'\sim P_n(v')}[c^{-}_{uv'} \nabla\mathcal{L}^-(\theta|u, v')] \bigg]\\
	=& \nabla \mathcal{L}(\theta^t)
	\end{split}
	\end{equation}
	The first equality is due to the definition of sampling procedure, and the second equality is due to the properties of joint probability distribution and expectation.
	
	\paragraph{Stratified Sampling with Negative Sharing}
	Under this setting, we follow a two-step sampling procedure: (1) draw an item $v\sim P_d(v)$, and (2) draw positive users $u \sim P_d(u|v)$. Negative links are constructed from independently drawn items in the same batch. So the batch $B$ consists of $m$ positive links and $n$ negative links. 
	
	We can use the same method as in Negative Sharing to calculate the probability of sampled negative links, which is also $(u, v) \sim P_d(u, v)$. Again, negative terms are re-weighted, thus we have
	\begin{equation}
	\begin{split}
	& \mathbb{E}_B[\nabla \mathcal{L}_{B}(\theta^t)] \\
	=& \frac{1}{m} \sum_{i=1}^{m} \mathbb{E}_{v\sim P_d(v), u\sim P_d(u|v)}  [c^{+}_{uv}\nabla\mathcal{L}^+(\theta|u, v)] +\frac{1}{n}\sum_{j=1}^{n}\mathbb{E}_{(u, v)\sim P_d(u, v)} [c^{-}_{uv} \frac{P_n(v)}{P_d(v)} \nabla\mathcal{L}^-(\theta|u, v)] \\
	=& \mathbb{E}_{(u, v)\sim P_d(u, v)} [c^{+}_{uv}\nabla\mathcal{L}^+(\theta|u, v)] + \mathbb{E}_{(u, v')\sim P_d(u) P_n(v')}[c^{-}_{uv'} \nabla\mathcal{L}^-(\theta|u, v')] \\
	=& \mathbb{E}_{u\sim P_d(u)}\bigg[
	\mathbb{E}_{v\sim P_d(v|u)}[c^{+}_{uv}\nabla\mathcal{L}^+(\theta|u, v)] + \mathbb{E}_{v'\sim P_n(v')}[c^{-}_{uv'} \nabla\mathcal{L}^-(\theta|u, v')] \bigg]\\
	=& \nabla \mathcal{L}(\theta^t)
	\end{split}
	\end{equation}
	The first equality is due to the definition of sampling procedure, and the second, third and fourth equality is due to the properties of joint probability distribution and expectation.
\end{proof}

\begin{proposition0}\label{th:convergence}
	Suppose $\mathcal{L}$ has $\sigma$-bounded gradient; let $\eta_t = \eta = c/ \sqrt{T}$ where $c = \sqrt{\frac{2(\mathcal{L}(\theta^0) - \mathcal{L}(\theta^*)}{L \sigma^2}}$, and $\theta^*$ is the minimizer to $\mathcal{L}$. Then, the following holds for the sampling strategies given in Algorithm 2, 3, 4, 5
	
	$$
	\min_{0\le t\le T-1} \mathbb{E}[\|\nabla \mathcal{L}(\theta^t)\|^2] \le  \sqrt{\frac{2(\mathcal{L}(\theta^0) - \mathcal{L}(\theta^*))}{T}} \sigma
	$$
\end{proposition0}

\begin{proof}
	With the property of $L$-$smooth$ function $\mathcal{L}$, we have
	\begin{equation}
	\label{eq:apply_smooth}
	\mathbb{E}[\mathcal{L}(\mathbf{\theta}^{t + 1})] \le \mathbb{E}[\mathcal{L}(\mathbf{\theta}^{t}) + \langle\nabla\mathcal{L}(\mathbf{\theta}^t), \mathbf{\theta}^{t + 1} - \mathbf{\theta}^{t}\rangle + \frac{L}{2}\|\mathbf{\theta}^{t+1} - \mathbf{\theta}^t\|^2]
	\end{equation}
	
	By applying the stochastic update equation, lemma \ref{th:convergence_lemma}, i.e. $\mathbb{E}_B[\nabla \mathcal{L}_{B}(\theta^t)] = \nabla\mathcal{L}(\theta^t)$, we have
	\begin{equation}
	\label{eq:apply_lemma}
	\begin{split}
	&\mathbb{E}[\langle\nabla\mathcal{L}(\mathbf{\theta}^t), \mathbf{\theta}^{t + 1} - \mathbf{\theta}^{t}\rangle + \frac{L}{2}\|\mathbf{\theta}^{t+1} - \mathbf{\theta}^t\|^2] \\
	\le& \eta_t \mathbb{E}[\|\nabla\mathcal{L}(\mathbf{\theta}^t)\|^2] + \frac{L\eta_t^2}{2}\mathbb{E}[\|\nabla\mathcal{L}_B(\mathbf{\theta}^t)\|^2]
	\end{split}
	\end{equation}

	Combining results in Eq. \ref{eq:apply_smooth} and \ref{eq:apply_lemma}, with assumption that the function $\mathcal{L}$ is $\sigma$-$bounded$, we have
	\begin{equation*}
	\mathbb{E}[\mathcal{L}(\mathbf{\theta}^{t + 1})] \le \mathbb{E}[\mathcal{L}(\mathbf{\theta}^{t})] + \eta_t \mathbb{E}[\|\nabla\mathcal{L}(\mathbf{\theta}^t)\|^2] + \frac{L\eta_t^2}{2b}\sigma^2
	\end{equation*}
	Rearranging the above equation we obtain
	\begin{equation}
	\label{eq:recursive_bound}
	\mathbb{E}[\|\nabla\mathcal{L}(\mathbf{\theta}^t)\|^2] \le \frac{1}{\eta_t}\mathbb{E}[\mathcal{L}(\mathbf{\theta}^{t} - \mathcal{L}(\mathbf{\theta}^{t + 1})] + \frac{L\eta_t}{2b}\sigma^2
	\end{equation}
	By summing Eq. \ref{eq:recursive_bound} from $t=0$ to $T - 1$ and setting $\eta= c/ \sqrt{T}$, we have
	\begin{equation}
	\begin{split}
	\min_{t} \mathbb{E}[\|\nabla \mathcal{L}(\mathbf{\theta}^t)\|^2] \le & \frac{1}{T} \sum_{0}^{T-1} \mathbb{E}[\|\mathcal{L}(\mathbf{\theta}^t)\|^2] \\
	\le & \frac{1}{c \sqrt{T}}(\mathcal{L}(\mathbf{\theta}^0) - \mathcal{L}(\mathbf{\theta}^*)) + \frac{Lc}{2\sqrt{T}} \sigma^2
	\end{split}
	\end{equation}
	By setting
	$$
	c = \sqrt{\frac{2(\mathcal{L}(\mathbf{\theta}^0) - \mathcal{L}(\mathbf{\theta}^*))}{L\sigma^2}}
	$$
	We obtain the desired result.
\end{proof}
\section{Vector Dot Product Versus Matrix multiplication}

\begin{figure}[H]
	\centering
	\includegraphics[width=0.5\textwidth]{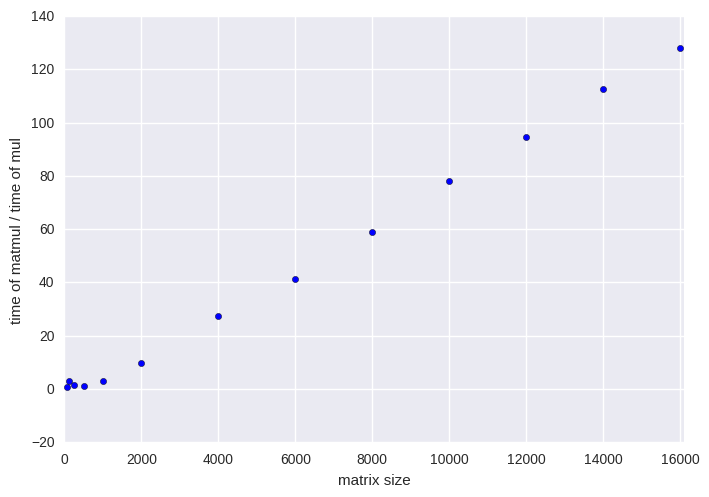}
	\caption{The computation time ratio between matrix multiplication and element-wise matrix multiplication for different square matrix sizes.}
	\label{fig:matmul_vs_mul}
\end{figure}

Here we provide some empirical evidence for the computation time difference of replacing vector dot product with matrix multiplication. Since vector dot product can be batched by element-wise matrix multiplication followed by summing over each row. We compare two operations between two square matrices of size $n$: (1) element-wise matrix multiplication, and (2) matrix multiplication. A straightforward implementation of the former has algorithmic complexity of $O(n^2)$, while the latter has $O(n^3)$. However, modern computation devices such as GPUs are better optimized for the latter, so when the matrix size is relatively small, their computation time can be quite similar. This is demonstrated in Figure \ref{fig:matmul_vs_mul}. In our choice of batch size and embedding dimension, $n \ll 1000$, so the computation time is comparable. Furthermore, $t_i\ll t_g$, so even several times increase would also be ignorable.

\section{Functional Embedding Versus Functional Regularization}

In this work we propose a functional embedding framework, in which the embedding of a user/item is obtained by some function such as neural networks. We notice another approach is to penalize the distance between user/item embedding and the function output (instead of equate them directly as in functional embedding), which we refer as functional regularization, and it is used in \cite{wang2015collaborative}. More specifically, functional regularization emits following form of loss function:
$$
\mathcal{L}(\mathbf{h}_u, \mathbf{h}_v) + \lambda \|\mathbf{h}_u - \mathbf{f}(\mathbf{x}_u)\|^2
$$

Here we point out its main issue, which does not appear in Functional Embedding. In order to equate the two embedding vectors, we need to increase $\lambda$. However, setting large $\lambda$ will slow down the training progress under coordinate descent. The gradient w.r.t. $\mathbf{h}_u$ is $\nabla_{\mathbf{h}_u} \mathcal{L}(\mathbf{h}_u, \mathbf{h}_v) + \frac{\lambda}{2} (\mathbf{h}_u - \mathbf{f}(\mathbf{x}_u))$, so when $\lambda$ is large, $\mathbf{h}^{t+1}_u \rightarrow \mathbf{f}^{t}(\mathbf{x}_u)$, which means $\mathbf{h}_u$ cannot be effectively updated by interaction information.


\end{document}